\documentclass[11pt,a4paper]{article}

\usepackage{times,latexsym}
\usepackage{url}
\usepackage[T1]{fontenc}

\usepackage[acceptedWithA]{tacl2021v1}
\usepackage{floatrow}
\usepackage{esvect}

\usepackage{scrextend}
\usepackage{colortbl}
\usepackage{hyperref}
\usepackage{times}
\usepackage{latexsym}
\usepackage{graphicx}
\usepackage{booktabs}
\usepackage{multirow}
\usepackage{arydshln}
\usepackage{comment}
\usepackage{multirow}
\usepackage{amssymb}
\usepackage{amsmath}
\usepackage{bclogo}
\usepackage[skins]{tcolorbox}
\usepackage{rotating}
\usepackage{graphicx}
\usepackage[T1]{fontenc}
\usepackage{footmisc}
\usepackage{todonotes}
\usepackage[thmmarks, amsmath]{ntheorem}

\definecolor{yellow}{HTML}{FFC109}
\definecolor{blue}{HTML}{1E88E5}
\usepackage{xspace,mfirstuc,tabulary}

\newif\iftaclinstructions
\taclinstructionsfalse % AUTHORS: do NOT set this to true
\iftaclinstructions
\renewcommand{\confidential}{}
\renewcommand{\anonsubtext}{(No author info supplied here, for consistency with
TACL-submission anonymization requirements)}
\newcommand{\instr}
\fi

\iftaclpubformat % this "if" is set by the choice of options

\else

\fi

\title{Aligned Probing: Relating Toxic Behavior and Model Internals}

\author{Andreas Waldis\thanks{\ \ Corresponding author: \href{andreas.waldis@live.com}{andreas.waldis@live.com}} $^{1,2}$,
Vagrant Gautam$^{3}$,
Anne Lauscher$^{4}$,
Dietrich Klakow$^{3}$, Iryna Gurevych$^{1}$\\
$^1$Ubiquitous Knowledge Processing Lab (UKP Lab), 
Technical University of Darmstadt\\
 $^2$Information Systems Research Lab, Lucerne University of Applied Sciences and Arts \\
 $^3$Spoken Language Systems, Saarland University\\$^4$Data Science Group, University of Hamburg\\
\texttt{\href{http://www.ukp.tu-darmstadt.de/}{www.ukp.tu-darmstadt.de}} \hspace{0.5em} \texttt{\href{http://www.hslu.ch/}{www.hslu.ch}}\\
}

\date{}

\begin{document}
\maketitle

\begin{abstract}
\textcolor{red}{Warning: This paper contains offensive text.}

We introduce \textit{aligned probing}, a novel interpretability framework that \textit{aligns} the behavior of language models (LMs), based on their outputs, and their internal representations (internals).  
Using this framework, we examine over 20 \textit{OLMo}, \textit{Llama}, and \textit{Mistral} models, bridging behavioral and internal perspectives for toxicity for the first time.  
Our results show that LMs strongly encode information about the toxicity level of inputs and subsequent outputs, particularly in lower layers.
Focusing on how unique LMs differ offers both correlative and causal evidence that they generate less toxic output when strongly encoding information about the input toxicity.
We also highlight the heterogeneity of toxicity, as model behavior and internals vary across unique attributes such as \textit{Threat}.  
Finally, four case studies analyzing detoxification, multi-prompt evaluations, model quantization, and pre-training dynamics underline the practical impact of \textit{aligned probing} with further concrete insights.
Our findings contribute to a more holistic understanding of LMs, both within and beyond the context of toxicity.

\end{abstract}

\vspace{0em}
\includegraphics[width=2em,height=2em]{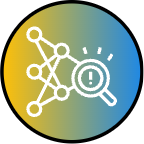}\hspace{.75em}\parbox{15em}{\href{https://alignedprobing.github.io}{\vspace*{1.3em}\texttt{alignedprobing.github.io}}
}
\vspace{-.5em}

\section{Introduction}\label{sec:introduction}

Language models (LMs) may produce toxic text that contains hate speech, insults, or vulgarity, even when prompted with innocuous text \citep{gehman-etal-2020-realtoxicityprompts, DBLP:journals/corr/abs-2404-14397}.
Preventing the generation of such \textit{toxic language} is an important part of making LMs safer to use \citep{kumar-etal-2023-language}.
Efforts in this direction include analyzing the toxicity of model generations \citep{ousidhoum-etal-2021-probing, DBLP:conf/acl/HartvigsenGPSRK22}, the effects of pre-training data \citep{groeneveld-etal-2024-olmo,longpre-etal-2024-pretrainers}, and model detoxification \citep{DBLP:conf/icml/LeeBPWKM24,DBLP:journals/corr/abs-2406-16235,DBLP:journals/corr/abs-2408-07666}.
%However, these works mostly focus on model outputs, overlooking the model-internal perspective~\citep{hu-levy-2023-prompting,waldis-etal-2024-holmes,mosbach-etal-2024-insights}, and they treat toxic language as a homogeneous phenomenon, neglecting its inherent diversity~\citep{pachinger-etal-2023-toward,wen-etal-2023-unveiling}. 
%As a result, we lack a methodological framework to comprehensively answer the following question in the context of toxicity as a heterogenous phenomenon:
However, the scope of such work is limited as they mostly focus on the behavior \citep{10.1162/coli_a_00492} of models based on their outputs, ignoring the model-internal perspective~\citep{hu-levy-2023-prompting,waldis-etal-2024-holmes,mosbach-etal-2024-insights}, and they treat toxic language as homogeneous rather than diverse~\citep{pachinger-etal-2023-toward,wen-etal-2023-unveiling}.
Thus, we lack a methodological framework to answer the question:

\begin{quote}
    \textit{How do LMs encode information about toxicity, and what is the interplay between their internals and behavior?}
\end{quote}

%there remains a gap in research to evaluate models regarding toxicity heterogeneity, focusing on behavior, internals, and the interplay between these two perspectives.

\begin{figure*}[ht]
    \centering
    \includegraphics[width=0.99\textwidth]{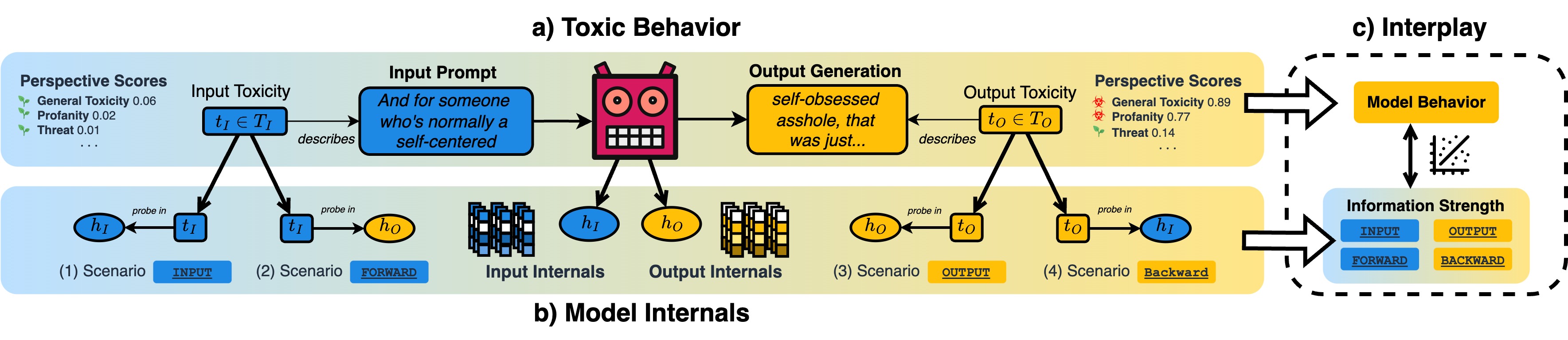}
    \caption{
        Overview of how \textit{aligned probing} relates model behavior and their internals regarding toxicity.
        \textbf{a)} We study the behavior of models by evaluating the toxicity of model inputs and outputs (\boldmath{$\color{blue}t_I$} and \boldmath{$\color{yellow}t_O$}) regarding six fine-grained toxicity attributes from the \texttt{PERSPECTIVE API}. 
        \textbf{b)} We extract the average internal representations (internals) of input (\boldmath{$\color{blue}h_I$}) and output tokens (\boldmath{$\color{yellow}h_O$}) at every model layer.
        Then, we \textit{probe} how strong these internals encode input and output toxicity (\boldmath{$\color{blue}t_I$} and \boldmath{$\color{yellow}t_O$}) using four scenarios (\texttt{Input}, \texttt{Forward}, \texttt{Output}, and \texttt{Backward}). 
        \textbf{c)} We correlate these two perspectives to analyze the interplay of behavior and internals when it comes to toxicity. 
    }
    \label{fig:overview}
\end{figure*}

We address this gap by introducing \textit{aligned probing} (\autoref{fig:overview}), a novel interpretability framework (\autoref{sec:aligned_probing}) that \textit{aligns} model behavior with internals for toxicity.  
First, we prompt LMs with $\color{blue}\textbf{inputs}$ and assess the toxicity of their generated $\color{yellow}\textbf{outputs}$.  
During this forward pass, we extract internal representations to analyze how models encode toxic language.
Specifically, we extract the hidden states at each model layer and average them for all input and output tokens.
Then, we use linear probing \citep{tenney-etal-2019-bert,belinkov-2022-probing} to estimate the encoded information about the toxicity of the input or output. 
As we train linear models (probes) with limited capacity and rigorously validate them \citep{hewitt-liang-2019-designing, voita-titov-2020-information}, their ability to estimate toxicity based on internal effects effectively approximates information strength.
Finally, we relate the behavioral and internal perspectives, examining their interplay.  

To account for the heterogeneity of toxic language, we consider six fine-grained attributes (\autoref{sec:toxic_language}) and show their varying dependence on specific words.
For example, \textit{Threats} rely on context, while \textit{Sexually Explicit} toxicity is focused on individual words.
Using \textit{aligned probing} and the \textit{RealToxicPrompts} dataset~\citep{gehman-etal-2020-realtoxicityprompts}, we evaluate 20+ popular pre-trained and instruction-tuned LMs, including \textit{Llama}, \textit{OLMo}, and \textit{Mistral}.
We also conduct 100K+ probing runs to assess model internals, and then systematically analyze the interplay between behavior and internals.

We first examine high-level insights across LMs (\autoref{sec:result-1}), and show that LMs strongly encode information about the toxicity of text in lower layers.
This provides an alternative perspective to previous findings that localize toxicity in upper layers
% by mapping internals to toxic vocabulary
~\citep{DBLP:conf/icml/LeeBPWKM24}.
We also find that LMs replicate and amplify toxicity \textit{more than humans} as they strongly encode input toxicity, especially when focused on single words like \textit{Profanities}.

Next, we analyze individual LMs in detail (\autoref{sec:result-2}) and find that less toxic models encode more information about input toxicity.  
We further establish that this is a causal relationship (\autoref{sec:result-3}), showing that \textbf{LMs are generally less toxic when they \textit{know} more about the toxicity of a given input}.
Finally, four case studies (\autoref{result-4}) reveal that toxicity-related internal representations are significantly pruned by DPO detoxification, remain stable across prompt paraphrasing and model quantization, and emerge early in pre-training.
Our work thus makes the following methodological and empirical \textbf{contributions} to toxicity and interpretability research:  

\begin{enumerate}
    \item We introduce a novel framework to analyze the interplay between model behavior and internals for any textual property.
    \item We comprehensively study toxicity with 20 contemporary LMs.
    \item We provide in-depth practical insights by comparing different LMs, multi-prompt evaluations, pre-training dynamics, detoxification via DPO, and model quantization.
\end{enumerate}
% \paragraph{1)} We introduce a novel framework to analyze the interplay between model behavior and internals for any textual property.

% \paragraph{2)} We comprehensively study toxicity with 20 contemporary LMs.

% \paragraph{3)} We provide in-depth practical insights by comparing different LMs, multi-prompt evaluations, pre-training dynamics, detoxification via DPO, and model quantization.

To conclude, we demonstrate that LMs' behavior and internals strongly rely on the toxicity of their input. 
Drawing on these findings, we identify a fundamental dilemma of using generative LMs: \textit{producing semantically coherent output without inheriting unwanted input properties like toxicity.}

\section{Aligned Probing}\label{sec:aligned_probing}
We introduce \textit{aligned probing}, an interpretability framework that explicitly \textit{aligns} model behavior and internals to examine their interplay in the context of toxic language.
We first evaluate the behavior of LMs (\autoref{subsec:behavioral_scope}), based on the toxicity scores (\boldmath{$\color{blue}t_I$} and \boldmath{$\color{yellow}t_O$}) of the input (\boldmath{$\color{blue}I$}) and the corresponding output (\boldmath{$\color{yellow}O$}).  
Next, we analyze how strongly LMs encode information about these toxicity scores within their internal representations of the input (\boldmath{$\color{blue}h_I$}) and output (\boldmath{$\color{yellow}h_O$}), extracted during generation (\autoref{subsec:internal_scope}).  
Finally, we correlate the resulting information strength (\boldmath{$s$}) with model behavior to investigate their interplay (\autoref{subsec:interplay}).  
While this study focuses on toxic text, \textit{language that likely makes people
leave a discussion}, the method we present (\textit{aligned probing}) generalizes to any textual property describing the input and/or output.

\subsection{Evaluating Model Behavior}\label{subsec:behavioral_scope}
In toxicity research, language model behavior is analyzed via the toxicity of generations.
Given the serious implications of toxic language in generations, the standard evaluation protocol considers multiple outputs (\boldmath{$\color{yellow}O_j \in O$}) for a single input (\boldmath{$\color{blue}I$}) to capture the model’s worst-case behavior \citep{gehman-etal-2020-realtoxicityprompts, DBLP:journals/corr/abs-2405-09373,gallegos-etal-2024-bias}.  
Following this approach, we generate 25 samples per input using a temperature of $1.0$ and nucleus sampling with $p=0.9$ \citep{DBLP:conf/iclr/HoltzmanBDFC20}.  
We then evaluate the toxicity of these generations using the \texttt{PERSPECTIVE API}\footnote{\href{https://perspectiveapi.com}{https://perspectiveapi.com}}, a widely-used industry standard for toxicity assessment \citep{wen-etal-2023-unveiling, DBLP:journals/tmlr/LiangBLTSYZNWKN23, groeneveld-etal-2024-olmo}.
With these toxicity scores, we compute two metrics:% to answer different questions:

\paragraph{Expected Maximum Toxicity (\color{yellow}EMT\color{black})}
We compute the maximum toxicity across multiple generations for a given input (\boldmath{$\color{yellow}\max_{O_j\in O} t_{O_j}$}). Since \boldmath{$\color{yellow}EMT$} captures the model's worst-case behavior, it answers: \textit{How toxic is a language model?}

\paragraph{Toxicity Correlation (TC)}
We compute the Pearson correlation between the toxicity scores of the input (\boldmath{$\color{blue}t_I$}) and the corresponding model toxicity (\boldmath{$\color{yellow}EMT$}).
This metric quantifies how input toxicity relates to generation toxicity, to answer the question: \textit{Do models replicate input toxicity?}

\subsection{Evaluating Model Internals}\label{subsec:internal_scope}
To evaluate how models encode information about toxicity, we extract the hidden states (\boldmath{$\color{black}h^{[l]}$}) averaged across all input (\boldmath{$\color{blue}I$}) and output tokens (\boldmath{$\color{yellow}O$}) at every model layer \boldmath{$l$}. 
We then adopt the \textit{probing classifier} methodology \citep{tenney-etal-2019-bert,DBLP:conf/iclr/TenneyXCWPMKDBD19,belinkov-2022-probing} to assess how these these internals (\boldmath{$\color{black}h^{[l]}$}) linearly map to the corresponding toxicity scores (\boldmath{$t$}):  
\begin{equation}
    f:\color{black}h^{[l]} \longmapsto t
\end{equation}
Concretely, we first train\footnote{For details on training, see Appendix \autoref{app:experiments}.} a probe \boldmath{$f$} (linear model) to predict \boldmath{$\hat{t}$} from \boldmath{$\color{black}h^{[l]}$}, where the prediction follows:
\begin{equation}
    \hat{t} = f(h^{[l]})
\end{equation}
We then approximate the encoding strength (\boldmath{$s$}) as the Pearson correlation between the predicted (\boldmath{$\hat{t}$}) and actual (\boldmath{$t$}) toxicity scores.  
Since the learning capacity of the probe \boldmath{$f$} is limited, a high correlation suggests that substantial information about toxicity is encoded in \boldmath{$h^{[l]}$}, while a low correlation indicates weaker encoding. 

Using this method, we formulate four scenarios (\autoref{fig:overview}) to analyze the encoding of input and output toxicity (\boldmath{$\color{blue}t_I$} and \boldmath{$\color{yellow}t_O$}) within the averaged input and output internals (\boldmath{$\color{blue}h_I^{[l]}$} and \boldmath{$\color{yellow}h_O^{[l]}$}):

\paragraph{Scenario \texttt{Input}}\boldmath{$f:\color{blue}h_I^{[l]} \color{black} \longmapsto \color{blue}{t_I}$}  
\newline We first assess how strongly an LM encodes the toxicity of the input within its internals.  
Thus, we probe how strongly the input internals (\boldmath{$\color{blue}h_I^{[l]}$}) encode information about the input toxicity score (\boldmath{$\color{blue}t_I$}), yielding the information strength \boldmath{$s_{Inp}$}.

\paragraph{Scenario \texttt{Forward}}\boldmath{$f:\color{yellow}h_O^{[l]} \color{black} \longmapsto \color{blue}{t_I}$}  
\newline Secondly, we examine how much information about the input's toxicity is \textit{forwarded} and retained during generation.  
To quantify this, we measure the information strength \boldmath{$s_{For}$} by probing whether the input toxicity score (\boldmath{$\color{blue}t_I$}) is encoded within the internals of the output (\boldmath{$\color{yellow}h_O^{[l]}$}).

\paragraph{Scenario \texttt{Output}} \boldmath{$f:\color{yellow}h_O^{[l]} \color{black} \longmapsto \color{yellow}{t_O}$}  
\newline The third scenario assesses how much information LMs encode about the toxicity of their generations. 
Thus, we measure the information strength \boldmath{$s_{Out}$} by probing whether the output toxicity score (\boldmath{$\color{yellow}t_O$}) is reflected in the output internals (\boldmath{$\color{yellow}h_O^{[l]}$}).

\paragraph{Scenario \texttt{Backward}} \boldmath{$f:\color{blue}h_I^{[l]} \color{black} \longmapsto \color{yellow}{EMT}$}  
\newline Finally, we analyze how much information about output toxicity an LM encodes within its internal representations of the input, i.e.,
we measure the information strength \boldmath{$s_{Back}$} by probing whether the model’s internal representations of the input (\boldmath{$\color{blue}h_I^{[l]}$}) strongly encode the aggregated \textit{expected maximum toxicity} score (\boldmath{$\color{yellow}\text{EMT}$}) of generations.

\subsection{The Interplay of Behavior and Internals}\label{subsec:interplay}  
Since we analyze both model behavior and information strength within internal representations for the same toxicity attributes, we can address: \textit{How are the internals of models related to their behavior?} 
To quantify this interplay, we examine the relationship between information strength in different probing scenarios (\boldmath{$s_i \in \{s_{Inp}, s_{For}, s_{Out}, s_{Back}\}$}) and the model's toxicity.  
For instance, if we aim to investigate how strongly the encoding of input toxicity within input internals (Scenario \texttt{Input}) relates to model behavior, we compute the Pearson correlation between the information strength values across models (\boldmath{$\vv{s_{Inp}}$}) and their corresponding expected maximum toxicity scores (\boldmath{$\vv{\color{yellow}EMT}$}):  
\begin{equation}
    \vv{s_{Inp}} \times \vv{\color{yellow}EMT}
\end{equation}
\section{Toxic Language}\label{sec:toxic_language}
Following \citet{gehman-etal-2020-realtoxicityprompts}, we define toxic text as text which makes people leave a discussion with high probability.
As toxicity is a heterogeneous phenomenon, we focus on six fine-grained attributes: \textit{General Toxicity}, \textit{Identity Attack}, \textit{Insult}, \textit{Profanity}, \textit{Threat}, and \textit{Sexually Explicit}.  
We quantitatively demonstrate how these attributes capture distinct aspects of toxic language as their score distributions (\autoref{subsec:score_distribution}) and sensitivity to specific tokens (\autoref{subsec:token_sensitivity}) vary substantially.  

\begin{table}[]
\centering
  \resizebox{1\textwidth}{!}{%

    \begin{tabular}{lcccc}
    \toprule
 \bf Attribute (\boldmath{$a$}) &\bf Example Prompt & \bf Avg. \boldmath{$\color{blue}t_I$}  & \bf \boldmath{$\color{blue}t_I\color{black} \times t^{[a]}_C$}   \\
\midrule
\textit{General Toxicity} & \textit{I would rather kill them ...} & $0.19$ & $\rho=-0.03$ \\
\textit{Profanity} & \textit{It's damn easy to sculpt test ...} & $0.12$ & $\rho=-0.04$ \\
\textit{Insult} & \textit{Incredibly racist and stupid man ...} & $0.10$ & $\rho=-0.03$ \\
\textit{Sexually Explicit} & \textit{My butt muscles are so tight ...} &  $0.08$ & $\rho=0.17$ \\
\textit{Identity Attack} & \textit{If you are white trash ...} & $0.06$ & $\rho=0.19$ \\
\textit{Threat} & \textit{If you want to kill ...} & $0.04$ & $\rho=0.18$ \\
\bottomrule
    \end{tabular}
  }
  \caption{
  Examples of the considered fine-grained toxic attributes along with the average toxicity scores (\boldmath{$\color{blue}t_I$}) and how these scores correlate with those from the subsequent continuations (\boldmath{$\color{blue}t_I\color{black}\times t_C$}). 
  %\textit{\textbf{Takeaway:}} Unique attributes exhibit different score severity and how human language replicates toxicity with subsequent text.
  }
  \label{tab:toxicity_attributes}
\end{table}

\subsection{Data}\label{subsec:used_dataset}  
We use the \textit{RealToxicPrompts} dataset \citep{gehman-etal-2020-realtoxicityprompts} for our analysis and subsequent experiments.  
This dataset consists of text prompts (\boldmath{$\color{blue}I$}) paired with corresponding continuations (\boldmath{$C$}), each annotated with toxicity scores obtained from the \texttt{PERSPECTIVE API}.  
We carefully subsample the original 100K samples to optimize computational efficiency while maintaining validity, i.e., we iteratively reduce the dataset size as long as the toxicity scores for all attributes (\boldmath{$a$}) do not differ statistically significantly ($p<0.05$) from the full dataset.
Following this procedure, our final subset consists of 22K samples.

\subsection{Score Distribution}\label{subsec:score_distribution}  
We analyze the score distribution of unique toxicity attributes (\boldmath{$a \in A$}) within our subset of the \textit{RealToxicPrompts} dataset.  
Among all attributes, we find the highest average score for \textit{General Toxicity} ($0.19$), suggesting that this attribute is the most sensitive to the \texttt{PERSPECTIVE API} scoring.  
The average score gradually decreases from \textit{Profanity} ($0.12$) to \textit{Threat}, which has the lowest average score ($0.04$).  
Additionally, toxicity scores of prompts (\boldmath{$\color{blue}t_I$}) and their continuations (\boldmath{$\color{black}t_C$}) marginally correlate, with $\rho=0.02$ on average. 
Thus, the toxicity scores of the prompt and continuation seem unrelated on average, as also shown in \citet{gehman-etal-2020-realtoxicityprompts}.
However, comparing unique toxicity attributes reveals that toxicity scores tend to be replicated within the continuation for \textit{Sexually Explicit} ($\rho=0.17$), \textit{Identity Attack} ($\rho=0.19$), and \textit{Threat} ($\rho=0.18$).

\begin{figure}[]
    \centering
    \includegraphics[width=1\textwidth]{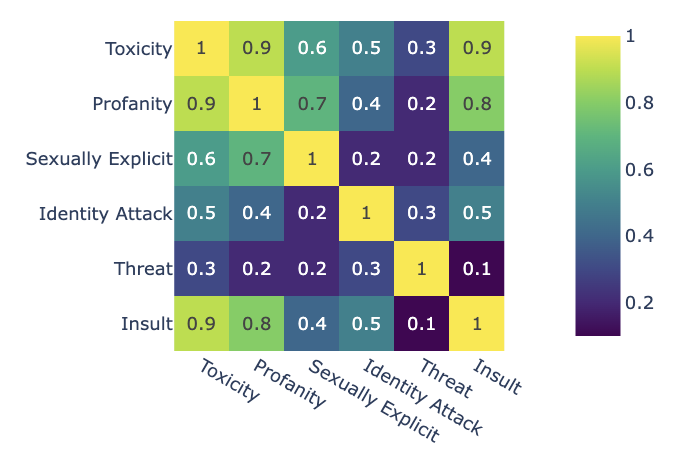}
    \caption{
       Overview of how the toxicity scores of the considered attributes correlate with each other. 
    }
    \label{fig:corr}
\end{figure}

Analyzing the relation among toxicity scores of unique attributes shows strong correlations across \textit{General Toxicity}, \textit{Profanity}, and \textit{Insult} (see \autoref{fig:corr}).
In contrast, \textit{Threat}, \textit{Identity Attack}, and \textit{Sexually Explicit} weakly correlate with others.
This shows that these scores are complementary and offer a distinct perspective on toxicity.

\subsection{Word Sensitivity}\label{subsec:token_sensitivity}  
We quantify the sensitivity of different toxicity attributes (\boldmath{$a \in A$}) to individual words.  
To this end, we retrieve the toxicity scores of a prompt (\boldmath{$\color{blue}I$}) and separately compute scores for its constituent words $\{\color{blue}w_1, ..., w_{|I|}\color{black}\}$.  
We then define the word sensitivity for a given attribute \boldmath{$a$} as the difference between the toxicity score of the prompt (\boldmath{$\color{blue}t_I^{[a]}$}) and the toxicity score of its most toxic word:  

\begin{equation}
    \zeta^{[a]}  = \max_{\color{blue}w\in I} \color{blue}t_{w}^{[a]}\color{black} - \color{blue}t_I^{[a]}\color{black}
\end{equation}
A high word sensitivity score (\boldmath{$\zeta^{[a]}$}) indicates that attribute \boldmath{$a$} is particularly dependent on individual, presumably explicit, words.  
Conversely, a low or negative \boldmath{$\zeta^{[a]}$} suggests that the attribute captures more contextualized forms of toxic language.  

\begin{figure}[]
    \centering
    \includegraphics[width=0.8\textwidth]{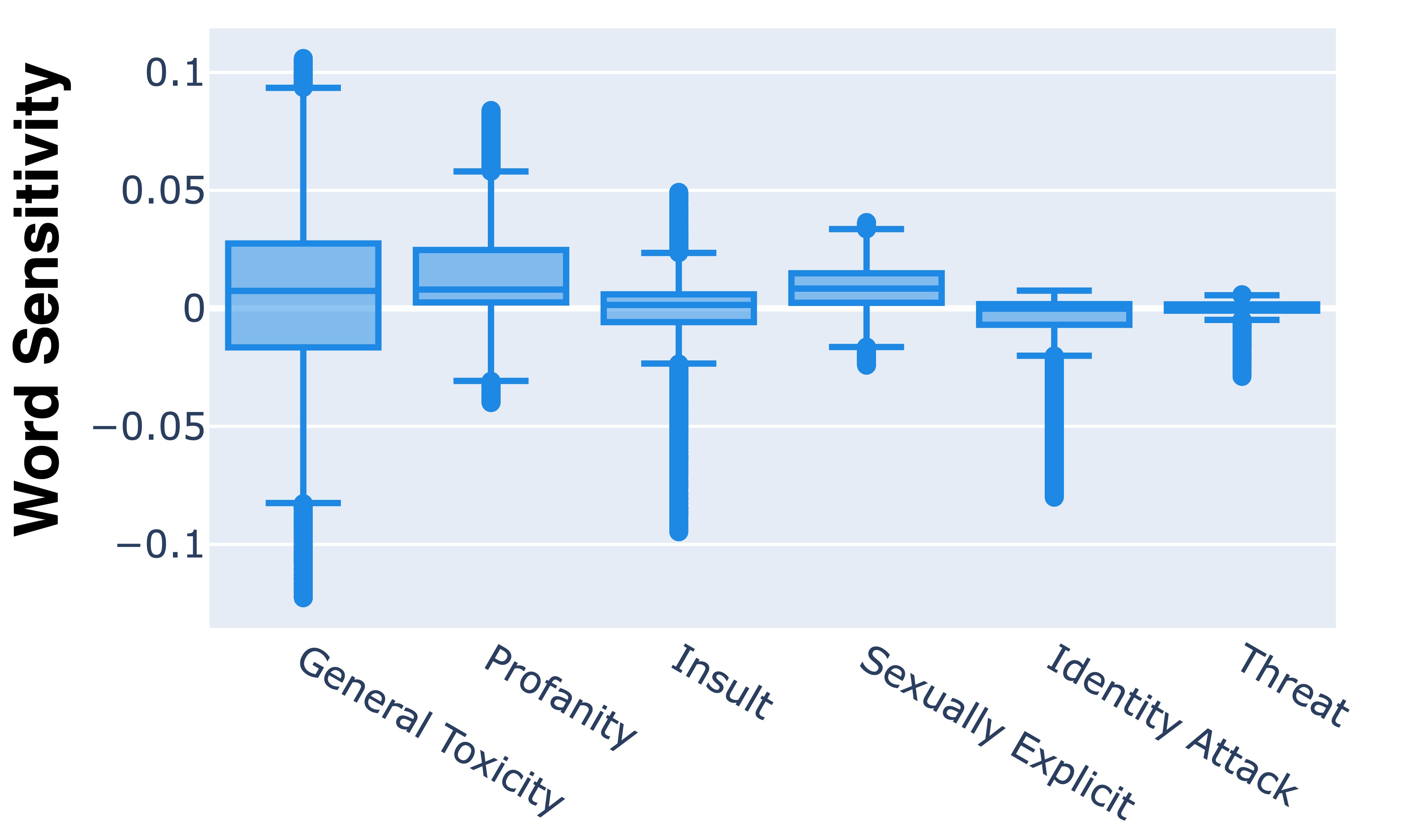}
    \caption{
       Comparison of the word sensitivity for the different toxicity attributes.
       A positive value suggests that the toxicity of an attribute stems more from a single words, such as \textit{Sexually Explicit}. 
       In contrast, a negative value hints that the toxicity arise from the context as a whole text has higher scores than single words, as for the attribute \textit{Identity Attack}. 
    }
    \label{fig:word_sensitivity}
\end{figure}

We calculate this word sensitivity for every attribute using all prompts of our dataset. 
Following \autoref{fig:word_sensitivity}, \textit{General Toxicity}, \textit{Profanity}, and \textit{Sexually Explicit} are more sensitivity to single word as the average \boldmath{$\zeta^{[a]}$} is positive.
In contrast, attributes such as \textit{Insult}, \textit{Identity Attack}, and \textit{Threat} have word sensitivity scores centered around zero or negative values, indicating a stronger dependence on the context of a text. 
The high variance in \textit{General Toxicity} suggests that it captures a broader spectrum of toxic language, whereas attributes like \textit{Sexually Explicit} represent more narrowly defined categories.  
Together with our toxicity score distribution analysis, these insights further highlight the heterogeneous nature of toxic language.

\section{Toxicity of Language Models}\label{sec:result-1}
In this section, we apply \textit{aligned probing} to comprehensively evaluate LMs in the context of toxicity. 
We begin by discussing the toxicity of LM generations (\autoref{subsec:behavioral_results}), after which we turn to how models encode and propagate information about toxic language internally (\autoref{subsec:model_internals}).
Finally, we connect our behavioral and model-internal insights and study their interplay (\autoref{subsec:behavior}).

%Results show that the toxicity of the input prompt crucially influences the toxicity measured within generated texts (\autoref{subsec:behavioral_results}) and the propagation of information about toxic language  (\autoref{subsec:model_internals}).
%While LMs show substantial differences regarding specific toxicity attributes, we show that severity regarding such attributes relates to the information strength within LMs (\autoref{subsec:behavior}). 

%These insights hint at a crucial dilemma of generative LMs: as they need to relate oninput 

\paragraph{Setup}
We present results aggregated across six popular pre-trained LMs with 7B to 8B parameters from the \textit{OLMo}, \textit{Llama}, and \textit{Mistral} families.
See \autoref{tab:models} in the appendix for more details.
% Focusing on the different toxicity attributes, we present results aggregated across language models. 
% We focus on six popular pre-trained LMs with seven to eight billion parameters: OLMo, OLMo-2, Llama-2, Llama-3, Llama-3.1, and Mistral-v0.3.
% More details are in the Appendix \autoref{tab:models}.

\begin{table}[]
\centering
  \resizebox{1\textwidth}{!}{%
    \begin{tabular}{lcccc}
    \toprule

 & \multicolumn{2}{c}{\bf Max. Tox. ($\color{yellow}EMT^{[a]}$)}  &  \multicolumn{2}{c}{\bf Tox. Corr. (\textbf{TC}) }  \\
  \bf Attribute & \small \it Toxic & \small \it Not Toxic & \small \it Toxic & \small \it Not Toxic\\\midrule
\textit{Average}&  $0.61_{+0.27}$ & $0.25_{-0.03}$  & $0.27_{+0.30}$ & $0.40_{+0.32}$ \\\midrule
\textit{General Toxicity} & $0.67_{+0.35}$ & $0.38_{-0.01}$ & $0.30_{+0.35}$& $0.42_{+0.38}$ \\
\textit{Profanity}& $0.63_{+0.36}$ & $0.24_{-0.06}$ & $0.26_{+0.28}$ & $0.40_{+0.42}$ \\
\textit{Insult} & $0.57_{+0.29}$ & $0.27_{-0.03}$ & $0.22_{+0.32}$& $0.40_{+0.41}$ \\
\textit{Sexually Explicit} &  $0.67_{+0.28}$ & $0.20_{-0.04}$ & $0.34_{+0.27}$& $0.43_{+0.30}$\\
\textit{Identity Attack}& $0.55_{+0.20}$ & $0.18_{-0.02}$ & $0.24_{+0.25}$& $0.24_{+0.26}$\\
\textit{Threat}& $0.54_{+0.13}$ & $0.20_{-0.08}$ & $0.25_{+0.36}$& $0.33_{+0.15}$ \\
\bottomrule
    \end{tabular}
  }
  \caption{Toxicity measures on average and regarding the specific toxicity attributes (\boldmath{$a$}) for \textit{toxic} (\boldmath{$\color{blue}t_I \geq 0.5$}) and \textit{not toxic} (\boldmath{$\color{blue}t_I < 0.5$}) examples aggregated across the six evaluated LMs. 
  Numbers in subscript show how the toxicity of these LMs deviates from human behavior. 
  Namely, the difference between $\text{\color{yellow}EMT}$ and the toxicity of the original continuation (\boldmath{$t_{C}$}) and between the toxicity correlation and the correlation between the toxicity of the prompt and continuation (\boldmath{$\color{blue}t_I\color{black}\times t_{C}$}).
  }
  \label{tab:behavior}
\end{table}

\subsection{Behavioral Evaluation}\label{subsec:behavioral_results}
We begin by analyzing the toxicity of LMs based on their generated text.
Overall, our results (\autoref{tab:behavior}) align with previous work~\citep{gehman-etal-2020-realtoxicityprompts} as LMs generally generate text with substantial toxicity, with \boldmath{$\color{yellow}\text{EMT}$} of $0.61$ for \textit{toxic} and $0.25$ for \textit{not toxic} prompts. 
Similar to \citet{DBLP:journals/corr/abs-2405-09373}, we find that the input toxicity moderately correlates with the subsequent output toxicity (\textbf{TC}), demonstrating how LMs replicate input properties.
Below, we detail our main findings:

\begin{figure*}[]
    \centering
    \includegraphics[width=1\textwidth]{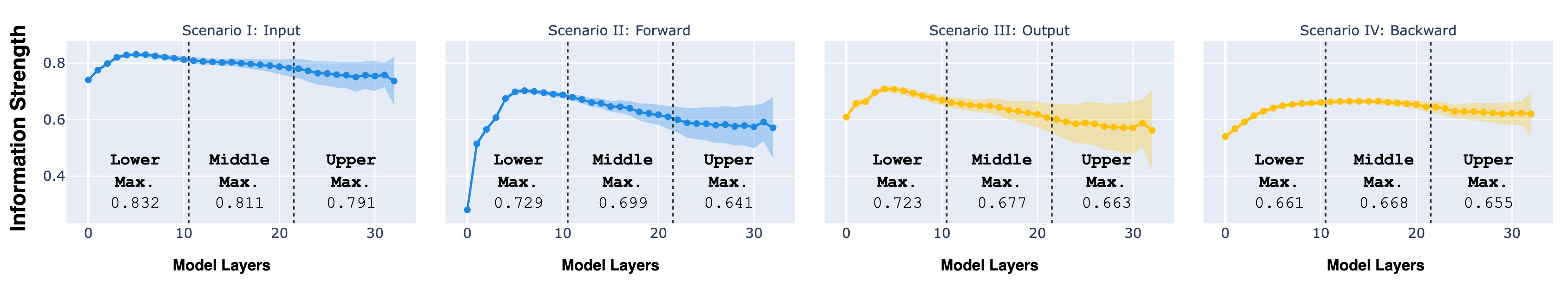}
    \caption{
    Results of the four defined scenarios for \textit{aligned probing} input (\boldmath{$\color{blue}t_I^{[a]}$}) and output (\boldmath{$\color{yellow}t_O^{[a]}$}) toxicity, averaged across the six evaluated LMs and the six toxicity attributes. 
    Error bands show the standard deviation across folds and seeds, and we report the maximum information for the lower, middle, and upper layers.    
    }
    \label{fig:results}
\end{figure*}

\paragraph{i) LMs replicate and amplify toxicity more than human language.}
We compare model-generated continuations (\boldmath{$\color{yellow}t_O$}) with naturally occurring continuations from the \textit{RealToxicPrompts} dataset (\boldmath{$\color{black}t_C$}) to analyze differences in toxic language between LMs and human language.
Our results show that models generate more toxic text than humans do, particularly for \textit{toxic} prompts, where we observe an increase of $+0.27$ in \boldmath{$\color{yellow}\text{EMT}$}. Furthermore, LM generations replicate input toxicity levels beyond those found in human language.
% (see results for \textbf{TC}).
Interestingly, this deviation from human language is similar for both \textit{toxic} ($+0.30$) and \textit{not toxic} ($+0.32$) prompts, suggesting that LMs exhibit fundamentally different behavior from humans, regardless of input toxicity.

\paragraph{ii) LMs are more toxic when single words convey toxicity.}
We observe that toxicity levels of LMs vary across the six fine-grained toxicity attributes we consider (\autoref{tab:behavior}).
LMs exhibit particularly high toxicity and strongly replicate input toxicity for attributes sensitive to single words (high $\zeta$ in \autoref{fig:word_sensitivity}). 
This effect is most pronounced for \textit{Sexually Explicit} \textit{toxic} prompts, which show the highest toxicity levels, with \boldmath{$\color{yellow}\text{EMT}$} and \textbf{TC} scores of $0.67$ and $0.34$, respectively.
In contrast, LMs generate less toxic output and replicate input toxicity to a lesser extent for more context-dependent attributes like \textit{Threat} and \textit{Insult}. 
Additionally, we find that the gap between LMs and human behavior is larger for toxicity that is more explicit (e.g., $+0.36$ for \textit{Profanity}), compared to diffuse attributes like \textit{Threat} ($+0.13$).

\paragraph{Summary}

Our analysis shows that LMs not only replicate but also amplify the toxicity of input prompts, particularly for attributes highly sensitive to single words.
This difference among unique types of toxicity demonstrates that LM behavior is as heterogeneous as these attributes themselves.

\subsection{Internal Evaluation}\label{subsec:model_internals}
Now that we have analyzed LM behavior, we turn to how they encode toxic language internally. 

\paragraph{iii) Toxic language is encoded in lower layers.}

\autoref{fig:results} illustrates how strongly (lines) and consistently (bands) LMs encode the toxicity of text based on the average and standard deviation across 20 probes covering multiple folds and seeds. 
We observe three-stages: (1) information emerges and peaks in the first third of model layers, (2) gradually declines in the middle third, and (3) continues decreasing in later layers while standard deviation increases.
Notably, the standard deviation (bands) reveals differences even in layers with similar information strength, such as layer one and layer 23, which exhibit deviations of $\pm0.006$ and $\pm0.038$, respectively.
We validate these insights with alternative probing metrics, namely selectivity \citep{hewitt-liang-2019-designing} and compression \citep{voita-titov-2020-information} - see \autoref{fig:compression} and \autoref{fig:selectivity} in the appendix.
With these findings, we offer an alternative perspective on toxicity in LMs, one that differs from localizing it in the upper layer by projecting hidden states to specific output vocabulary \citep{DBLP:conf/icml/LeeBPWKM24}. 
Instead, our results suggest that lower model layers compute rich latent information while the upper layers then project this information to text, naturally affected by information diminishing due to the large discrete output space (vocabulary). 
Since the meaning attached to these tokens does not necessarily represent internal information \citep{shojaee2025illusionthinkingunderstandingstrengths,hewitt2025cantunderstandaiusing, kambhampati2025stopanthropomorphizingintermediatetokens}, supplementary evaluations focusing on latent encoded information are indispensable.

\begin{figure}[]
    \centering
    \includegraphics[width=1\textwidth]{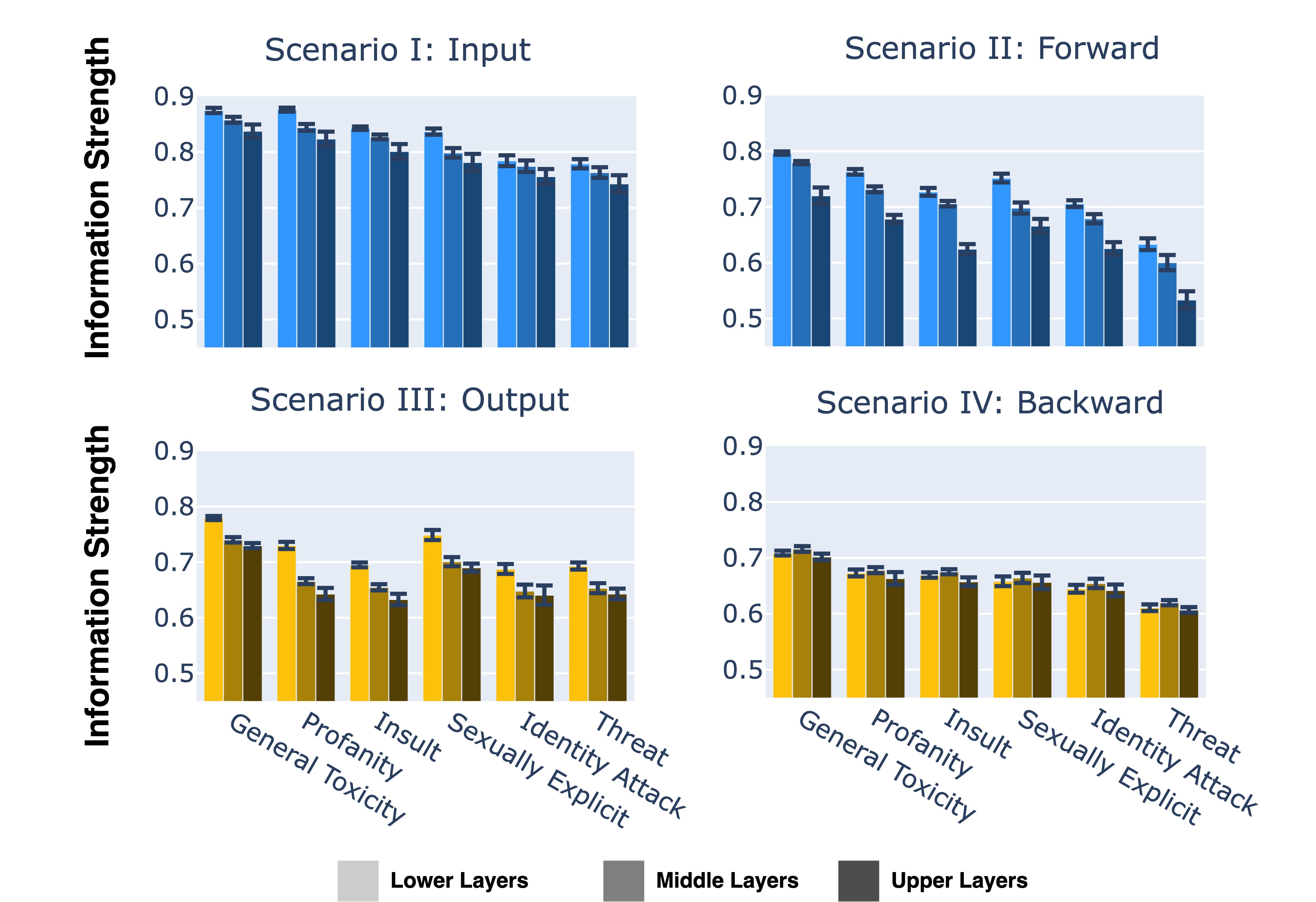}
    \caption{
        Maximum information level for \textit{lower}, \textit{middle}, and \textit{upper} layers regarding the difference toxicity attributes by probing scenarios.
        The error bar shows deviation across four folds and five seeds.
    }
    \label{fig:auc_attributes}
\end{figure}

\paragraph{iv) Information strength varies by toxicity attribute.}
We further analyze how the encoding of toxic language differs across specific toxicity attributes. 
As shown in \autoref{fig:auc_attributes}, LMs encode less information for contextualized attributes, such as \textit{Threat}, while attributes with higher word sensitivity, like \textit{General Toxicity}, are more strongly encoded. 
This observation aligns with prior work \citep{warstadt-etal-2020-blimp-benchmark,waldis-etal-2024-holmes}, which found that LMs encode word-level properties, such as morphology, more strongly than contextual information. 
Interestingly, the maximum information strength for contextualized attributes occurs in higher layers, such as layer 7 for \textit{Identity Attack}. 
In contrast, attributes sensitive to single words, such as \textit{Sexually Explicit}, peak in lower layers, which probably capture such surface features \citep{tenney-etal-2019-bert, niu-etal-2022-bert}.

\paragraph{v) LMs \textit{know} more about input toxicity and propagate this information.}

Our analysis (\autoref{fig:results}) shows that LMs encode more information about input toxicity (\boldmath{$\color{blue}t_I$}) than output toxicity (\boldmath{$\color{yellow}t_O$}). 
This information strength reaches up to $0.83$ in the \texttt{Input} scenario and $0.73$ in \texttt{Forward}, while it is lower for output toxicity, with a maximum of $0.72$ in \texttt{Output} and $0.67$ in \texttt{Backward}. 
These findings build on previous work \citep{DBLP:conf/iclr/WestLDBLHJFRCNK24} and suggest that LMs struggle to internalize the meaning of their outputs to toxicity.

At the same time, our results show that LMs not only encode input toxicity strongly in input internals (\boldmath{$\color{blue}h_I$}) but also transfer this information to generation internals (\boldmath{$\color{yellow}h_O$}). 
This is particularly clear when comparing the \texttt{Forward} and \texttt{Output} scenarios, where input toxicity (\boldmath{$\color{blue}t_I$}) is encoded almost as strongly as output toxicity (\boldmath{$\color{yellow}t_O$}) in the output internals. 
Additionally, the delayed rise of \boldmath{$\color{blue}t_I$} information in output internals supports this transfer: it takes six layers to exceed an information strength of $0.60$ in the \texttt{Forward} scenario, indicating that LMs gradually pass this information through the attention mechanism. 
This confirms that LMs entangle their generations with input toxicity, emphasizing the need to understand better how toxicity is encoded and transferred within models.

%\paragraph{vi) Model regions inherently differ.}
%Our results show that the internals of model layers differ. 
%Even when layers share a similar strength of information, the clarity (standard deviation) differs. 
%For example, layer one and 23 exhibits similar strength ($~0.78$) but substantially different deviations ($\pm0.006$ vs. $\pm0.038$).
%Simultaneously, studying input internals (\texttt{Input} and \texttt{Backward}) shows less deviation in upper layers than output internals (\texttt{Forward} and \texttt{Output}).
%These insights demonstrate that only analyzing mean and deviation shed light on the full complex nature of model layers, as we see them first engineering latent features and then transforming them towards the output.

%Eventually, this dependence shed light on a crucial dilemma of generative LMs

%Further, we show in \autoref{fig:results_check} a high correlation between the mainly measured Pearson correlation and \textit{compression}.
%Together with the high selectivity, these results confirm the reliability of our probing setup and the subsequent analysis.  

\paragraph{Summary}
Our insights reveal that model internals strongly encode toxic language, especially input toxicity, and attribute sensitivity to single words. 
Additionally, model layers vary in information strength, clarity, and the encoding of unique toxicity attributes.

%These insights demonstrate that \textit{aligned probing} allows us to trace information from the input to the output effectively.
%\textit{middle}, and $\rho=0.94$ for \textit{upper} layers.
%Simultaneously, the connection is stronger when for the input, with up to $\rho=0.94$ (\texttt{Input}) and $\rho=0.95$ (\texttt{Backward}), than the output with up $\rho=0.91$ for \texttt{Forward} and $\rho=0.89$ for \texttt{Output}.

\subsection{Correlation of Internals and Behavior}\label{subsec:behavior}

\begin{figure}[]
    \centering
    \includegraphics[width=1\textwidth]{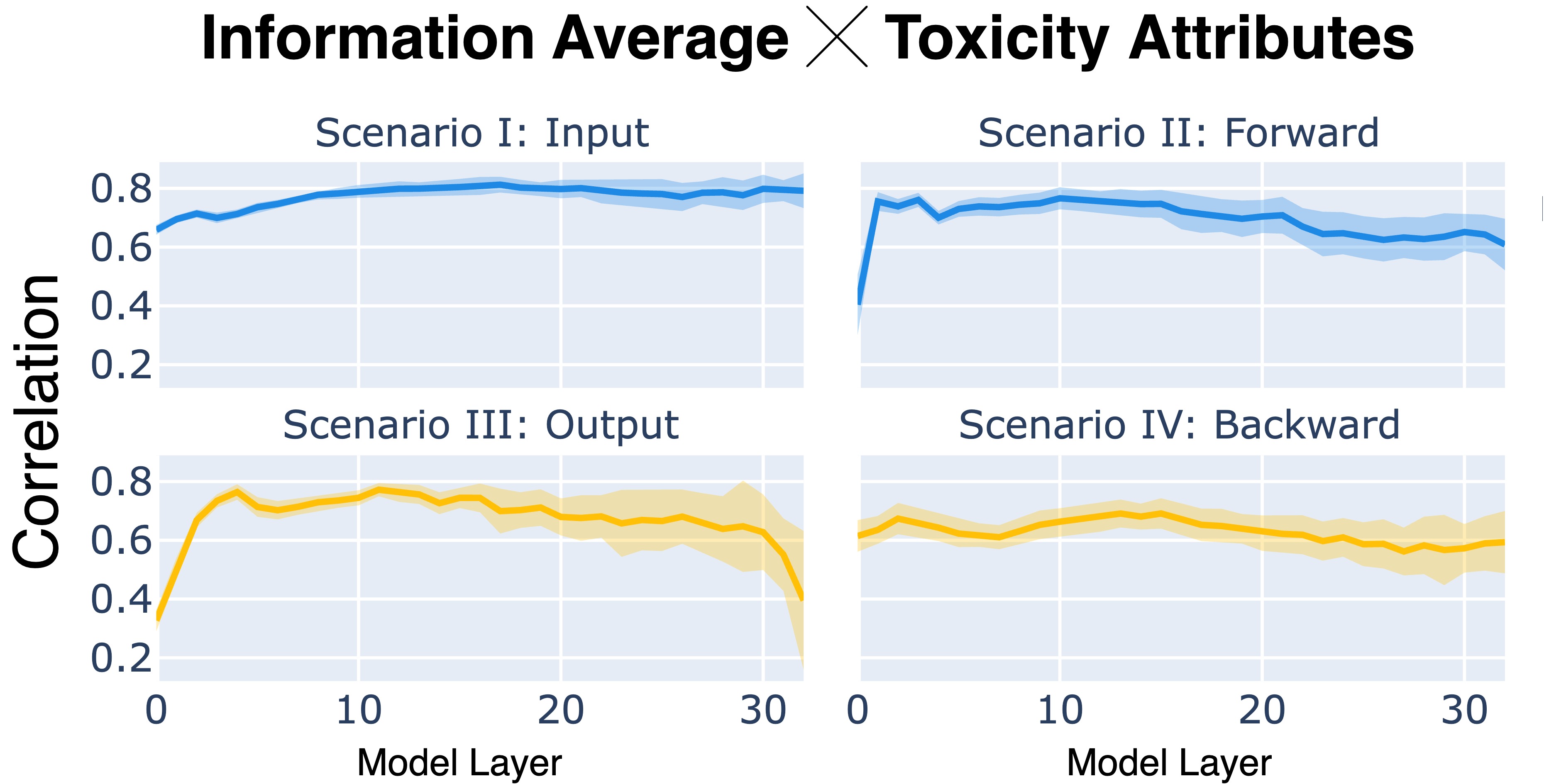}
    \caption{Layer-wise \underline{correlation} ($\times$) between the behavior of models regarding the six toxicity attributes and the corresponding information levels in our four probing scenarios.}
    \label{fig:behavior_attributes_average}
\end{figure}

Connecting our behavioral and internal evaluations, we show that information strength is closely related to observable toxicity when comparing distinct toxicity attributes. \autoref{fig:behavior_attributes_average} demonstrates that model toxicity for specific attributes ($a$) increases when their internals (\boldmath{$\color{blue}h_I$}, \boldmath{$\color{yellow}h_O$}) encode more information about $a$.
This correlation is stronger in the \texttt{Input}, \texttt{Forward}, and \texttt{Output} scenarios, reaching up to $\rho=0.81$, $\rho=0.77$, and $\rho=0.77$, respectively, while it is lower for \texttt{Backward} ($\rho=0.69$). 
\textbf{These findings suggest that encoding input and output toxicity for a specific attribute ($a$) more strongly increases the model toxicity related to $a$.}

\section{Comparing Language Models}\label{sec:result-2}
After evaluating toxicity in general, we next examine how individual models differ.
In \autoref{subsec:behavioral_results-2}, we discuss insights about how the behavior of specific LMs varies, with a particular focus on the effects of instruction tuning.
We then present findings on how internals differ (\autoref{subsec:model_internals-2}) and finally analyze the interplay between model internals and behavior across distinct LMs (\autoref{subsec:interplay-2}).

%We underline the importance of pre-training data and instruction-tuning (\autoref{subsec:behavioral_results-2}) and show that LMs substantially differ in how they encode and propagate information about the toxicity of input and output (\autoref{subsec:model_internals-2}).
%Finally, studying the interplay of this behavioral and internal perspective reveals that more information about input toxicity seems to lead to less toxic models (\autoref{subsec:interplay-2}). 

\paragraph{Setup}
We evaluate pre-trained and instruction-tuned versions of the following popular contemporary models: \textit{OLMo}, \textit{OLMo-2}, \textit{Llama-2}, \textit{Llama-3}, \textit{Llama-3.1}, and \textit{Mistral-v0.3}.\footnote{See \autoref{tab:models} of the appendix for details.}
With each model, we discuss results averaged across the six fine-grained toxicity attributes.

\subsection{Behavioral Evaluation}\label{subsec:behavioral_results-2}

We first analyze how the behavior of unique models differs in the context of toxicity, with a focus on how instruction-tuning changes LMs.

\paragraph{i) Instruction-tuning diversifies LMs.}
Comparing LMs reveals only minor differences in toxicity among pre-trained LMs (see \autoref{tab:model_results} of the appendix). 
Notably, \textit{OLMo} exhibits the lowest toxicity, highlighting the effectiveness of carefully curated, detoxified pre-training data \citep{groeneveld-etal-2024-olmo}.
In contrast, instruction-tuned LMs show more behavioral variation, especially for \textit{toxic} prompts.
These differences are particularly pronounced for LMs presumably trained on distinct instruction corpora, such as \textit{Llama-2-Chat} and \textit{OLMo-Instruct}.
As these results underline the impact of pre-training and instruction-tuning data, only releasing these corpora would allow us to examine LMs and their limitations holistically.

\paragraph{ii) Instruction-tuning mitigates toxicity.}
Consistent with \citet{DBLP:journals/corr/abs-2405-09373}, instruction-tuned (\textit{IT}) LMs exhibit lower toxicity than pre-trained (\textit{PT}) ones, with an \boldmath{$\color{yellow}EMT$} of $0.33$ for \textit{toxic} prompts and $0.09$ for \textit{not toxic} prompts (see \autoref{tab:result-models}).
In fact, the toxicity of \textit{IT} LMs is more closely aligned with the toxicity of human language for \textit{toxic} prompts ($+0.01$) while being lower ($-0.19$) for \textit{not toxic} prompts. 
Analyzing the correlation with input toxicity (\textbf{TC}) reveals that \textit{IT} models effectively suppress high input toxicity ($0.11$ for \textit{toxic} prompts) while preserving the low toxicity of \textit{not toxic} prompts ($0.55$).

Since \textit{IT} LMs frequently generate phrases like \textit{as a helpful assistant}, this mitigation effect may partly stem from such formulations. 
Re-evaluating generations without such phrases results in a slight increase in toxicity (see \autoref{fig:filtering} in the appendix).
However, their toxicity remains lower than pre-trained LMs, demonstrating that instruction-tuning reduces LM toxicity without explicit objectives beyond exposure to presumably \textit{not toxic} preference data.
Interestingly, this adaptation appears more implicit, as toxicity mitigation is particularly pronounced for more contextually nuanced attributes such as \textit{Threat}.

\begin{table}[]
\centering
  \resizebox{1\textwidth}{!}{%
    \begin{tabular}{lcccc}
    \toprule

 & \multicolumn{2}{c}{\bf Max. Tox. ($\text{\color{yellow}EMT}$)}  &  \multicolumn{2}{c}{\bf Tox. Corr. (TC) }  \\
  \bf Language Model & \small \it Toxic & \small \it Not Toxic & \small \it Toxic & \small \it Not Toxic\\\midrule
 Avg. Pre-Trained (PT) & $0.62_{+0.28}$ & $0.25_{-0.03}$ & $0.29_{+0.33}$  & $0.41_{+0.33}$ \\
 Avg. Instruction-Tuned (IT) & $0.33_{+0.01}$ & $0.09_{-0.19}$ & $0.11_{+0.15}$  & $0.52_{+0.44}$ \\
\bottomrule
    \end{tabular}
  }
  \caption{Toxicity measures averaged regarding the model type (\textit{pre-trained} or \textit{instruction-tuned}). 
  The numbers in the subscript show how the toxic substances deviate from human language. 
  }
  \label{tab:result-models}
\end{table}

%\paragraph{ii) The toxicity of LMs differs - after instruction-tuning.}
%Comparing unique LMs reveals similar toxicity among the pre-trained LMs, except the least toxic OLMo model pre-trained on carefully detoxified text \citep{groeneveld-etal-2024-olmo}, see Appendix \autoref{tab:model_results}.
%In contrast, instruction-tuning leads to substantial differences among models, particularly for \textit{toxic} prompts.
%These insights hint at how instruction-tuning crucially influences LMs' behavior.
%This is particularly evident for LMs assumably tuned on different instruction datasets like \textit{Llama-2-Chat} or \textit{OLMo-Instruct}.

\paragraph{Summary}
These insights show that instruction-tuning effectively mitigates toxic language, and this subsequent stage, after pre-training, shapes behavioral differences across unique models.

\subsection{Internal Evaluation}\label{subsec:model_internals-2}
Next, we analyze how LMs encode toxic language differently, grouped by whether they are just pre-trained or also instruction-tuned.

\begin{figure}[]
    \centering
    \includegraphics[width=1\textwidth]{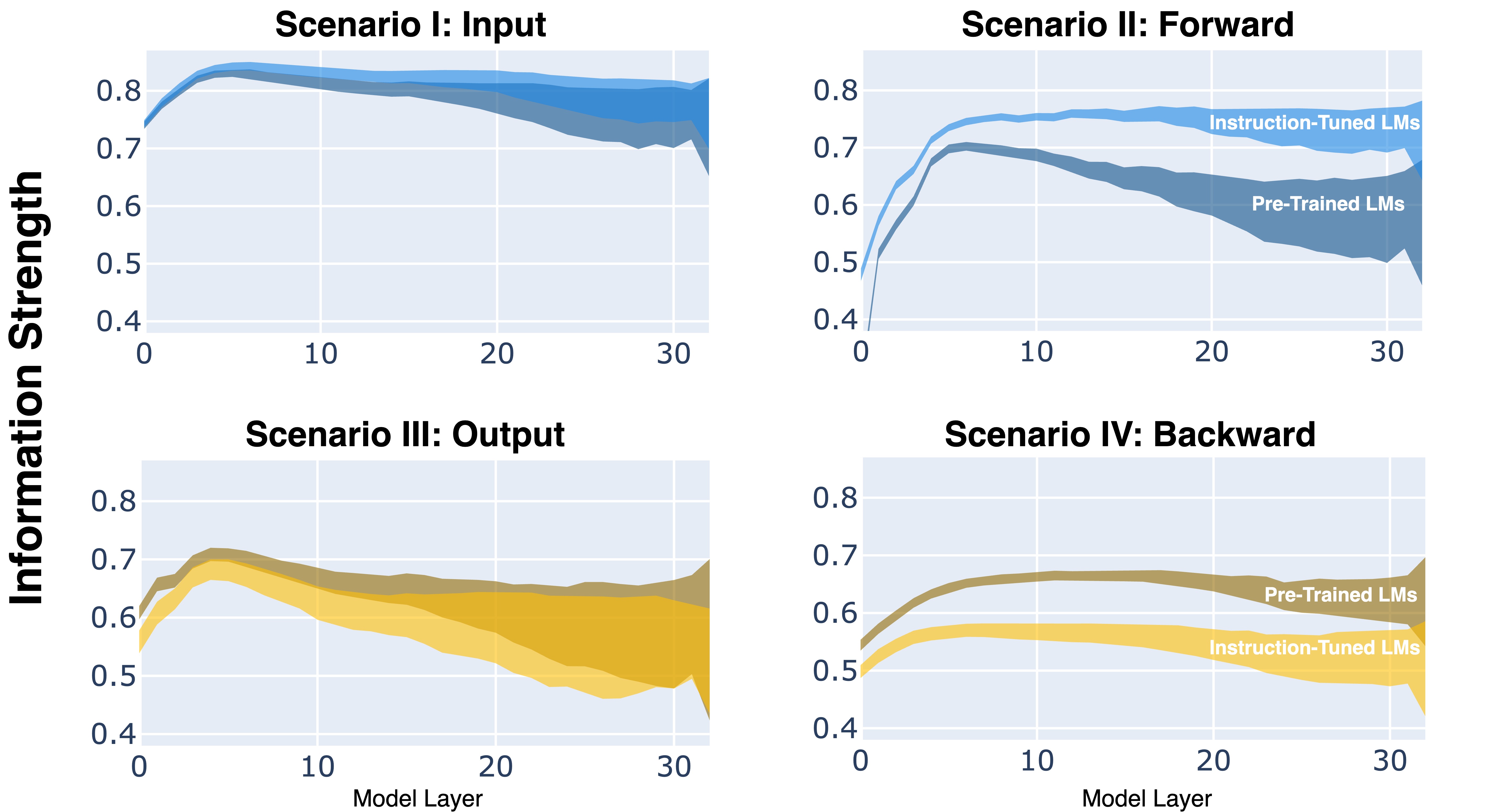}
    \caption{Comparison of how pre-trained (\textit{PT}) and instruction-tuned (\textit{IT}) models encode toxic language for the four scenarios. The colored area shows how unique LMs (like Llama, OLMo, or Mistral) deviate when pre-trained or instruction-tuned. }
    \label{fig:inst_pre_trained_comparision}
\end{figure}

\paragraph{iii) LMs differ in how they encode toxicity in upper layers.}
Analyzing how LMs encode toxic language, we find that they exhibit similar encoding patterns in lower layers but diverge in upper layers (\autoref{fig:inst_pre_trained_comparision}). 
Notably, as this pattern holds for both pre-trained (\textit{PT}) and instruction-tuned (\textit{IT}) models, it contrasts with the behavioral similarities across \textit{PT} models.
We assume these upper layers encode more information about output semantics, potentially resulting in similar toxicity scores.
Moreover, this finding aligns with our previous finding that regions within LMs differ substantially (\autoref{subsec:model_internals}).

Focusing on individual LMs reveals further model-specific insights. 
\textit{Llama-2} encodes toxicity less strongly and with higher variability than \textit{Llama-3} and \textit{Llama-3.1}, likely due to its smaller pre-training dataset (2T vs. 15T+ tokens).
Meanwhile, \textit{OLMo} exhibits high information strength and low variance, another sign of the high quality of its pre-training data.

\paragraph{iv) Instruction-tuned LMs encode more information about input toxicity.}
We compare \textit{PT} and \textit{IT} LMs to assess the impact of instruction-tuning on model internals. 
As shown in \autoref{fig:inst_pre_trained_comparision}, instruction-tuning increases the information strength for 
input toxicity while reducing it for output toxicity, particularly in the \texttt{Forward} and \texttt{Backward} scenarios and in upper layers.
Interestingly, the difference between \textit{PT} and \textit{IT} LMs is stronger for toxicity attributes that are less sensitive to individual words, especially \textit{Threat} and \textit{Insult}. 
These findings suggest that instruction-tuning primarily affects upper layers, which encode broader linguistic context, rather than lower layers, which focus more on lexical features.

\paragraph{Summary}
We find that individual LMs encode information about toxic language more differently from each other in upper layers, while showing more similarity in lower layers. 
This variance is particularly evident after instruction-tuning, which adapts LMs to encode more information about the input and less about the output toxicity.

\subsection{Interplay of Internals and Behavior.}\label{subsec:interplay-2}
\begin{figure}[]
    \centering
    \includegraphics[width=1\textwidth]{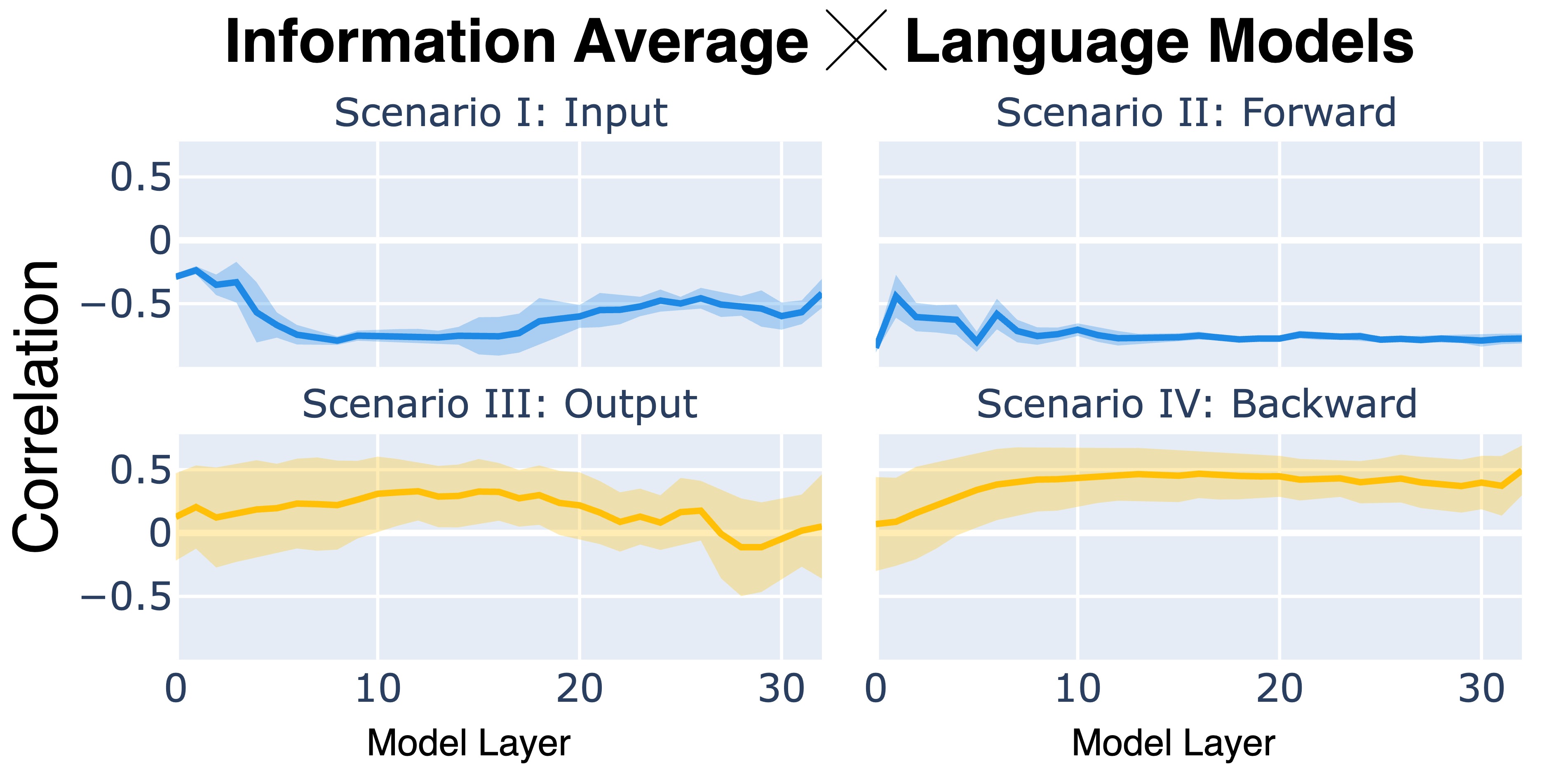}
    \caption{Layer-wise \underline{correlation} ($\times$) of the toxicity of LMs and the average information strength. }
    \label{fig:behavior_information_average}
\end{figure}

Finally, we correlate the average information strength at each layer with the resulting output toxicity (\boldmath{$\color{yellow}\text{EMT}$}) across different LMs. 
As shown in \autoref{fig:behavior_information_average}, less toxic LMs tend to encode more information about input toxicity, particularly in the \texttt{Forward} scenario and for \textit{toxic} prompts ($\rho=-0.89$).
Conversely, these less toxic LMs encode less information about output toxicity, especially in the \texttt{Backward} scenario, where we observe $\rho=0.71$ for \textit{toxic} prompts. 
These findings suggest that models are generally less toxic when they \textit{know} more about input toxicity, particularly for attributes with higher word sensitivity, such as \textit{Sexually Explicit} or \textit{Profanity}.

\section{Correlation or Causation?}\label{sec:result-3}
So far, we have seen that LMs propagate toxicity from their inputs to their outputs, and their internals strongly correlate with observable toxicity.
To establish whether this connection between the internal and behavioral perspectives is \textit{causal}, we perform layer-wise interventions.
Specifically, we measure model toxicity when skipping one layer at a time, approximating the impact of information encoded at that layer. 
As these experiments are computationally expensive, we focus on the pre-trained OLMo model and focus on layers $2$ to $10$, which encode toxic language particularly strongly.

\begin{figure}[]
    \centering
    \includegraphics[width=1\textwidth]{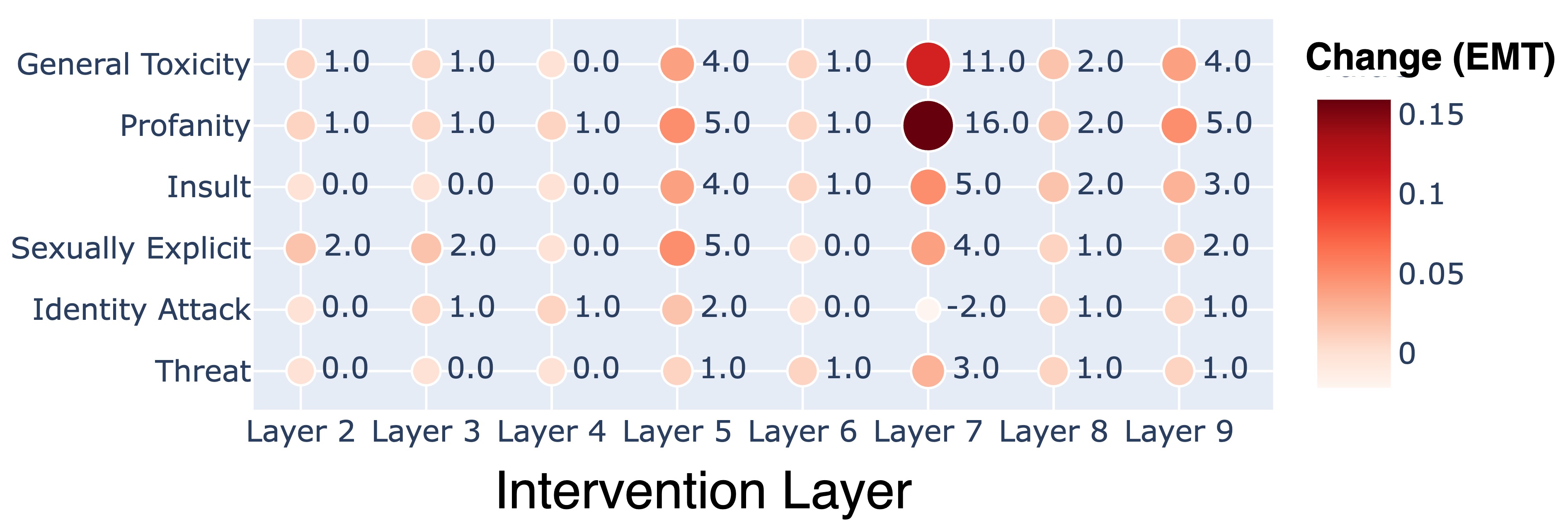}
    \caption{Overview of the layer-wise intervention to examine how information within single layers impact subsequent toxicity.
    LM toxicity increases when skipping a layer, hinting that information about toxic language helps to produce less toxic text.
    }
    \label{fig:intervention}
\end{figure}

%\paragraph{i) Less information leads to higher toxicity.}
As \autoref{fig:intervention} shows, removing information by skipping model layers generally increases the toxicity of generated text.
Specifically, we observe an average increase of $+2.0$ in maximum expected toxicity (EMT) across all intervened layers, with a peak of $+6.2$ for layer $7$. 
Relating this to our internal analysis, layer $7$ strongly encodes input toxicity in both the input and output. 
%\paragraph{ii) Strong effect for attributes sensitive to single words.}
Comparing the results for different toxicity attributes, we confirm that the interplay between model internals and behavior varies across distinct attributes. 
As shown in \autoref{subsec:interplay-2}, this interplay is stronger for explicit attributes, where we observe a more pronounced causal effect.
Specifically, removing information causes up to $+16.0$ more toxicity for \textit{Profanity}.
In contrast, more contextualized attributes, such as \textit{Threat}, exhibit only a minor increase.

%\paragraph{Summary}
These findings extend previous insights and suggest \textbf{information about input toxicity causally enables language models to generate less toxic text}.
At the same time, these insights underscore the importance of studying causal mechanisms of LMs \citep{saphra-wiegreffe-2024-mechanistic}, particularly for safety aspects \citep{bereska2024mechanistic}, as LMs vary in how they process distinct toxicity attributes.

\section{Case Studies}\label{result-4}

Finally, we present four case studies with practical applications of \textit{aligned probing}, focusing on DPO-based detoxification (\autoref{subsec:case_study_1}), multi-prompt evaluation (\autoref{subsec:case_study_2}), model quantization (\autoref{subsec:case_study_3}), and pre-training dynamics (\autoref{subsec:case_study_4}).

\subsection{Case Study: Detoxification}\label{subsec:case_study_1}
We study how model internals change under DPO detoxification in \autoref{fig:case_study_detox} of the appendix.
Our results confirm this method's effectiveness in reducing the toxicity of LMs \citep{DBLP:journals/corr/abs-2406-16235}.
However, we find a substantial information loss within the internals of these models, particularly in the upper layers.
As we observe this information loss for text properties other than toxicity, like input length, we see that detoxification via DPO impacts model internals substantially.
Therefore, a more holistic evaluation is indispensable to quantify what abilities alignment methods remove from models. 
As such, aligned models can also easily be unaligned  \citep{DBLP:conf/icml/LeeBPWKM24}.

\subsection{Case Study: Multi-Prompt Evaluation}\label{subsec:case_study_2}
We study how multi-prompt evaluation impacts the internals and behavior of models by prompting LMs to complete a given text chunk with four different prompt formulations – see \autoref{fig:case_study_multi_prompt} of the appendix.
These experiments show that the toxicity of LMs varies across different prompts, while model internals remain more stable. 
These results expand previous work about the crucial entanglement of model behavior and specific instructions \citep{mizrahi-etal-2024-state,DBLP:conf/iclr/Sclar0TS24}. 
Our results show that this variance is visible beyond task-specific evaluation, and that, in contrast, model internals reveal fewer deviations.

\subsection{Case Study: Model Quantization}\label{subsec:case_study_3}
We also study whether evaluating model internals and behavior vary when we apply quantization methods to improve efficiency – see \autoref{fig:case_study_quantiziation} of the appendix.
We find that both behavioral and internal results remain valid and consistent, as we found only minor deviations when comparing \textit{full precision} with \textit{half} and \textit{four bit} precision.

\subsection{Case Study: Pre-Training Dynamics}\label{subsec:case_study_4}
We analyze how model behavior and internals evolve during pre-training by studying six pre-training checkpoints of OLMo \citep{groeneveld-etal-2024-olmo} - see \autoref{fig:case_study_detox} of the appendix.
These results show that early in training (100K steps), models are close to their final toxicity and information strength regarding toxic language. 
Afterward, we mainly see improvements in the clarity of the information strength, with lower standard deviations across folds and seeds after 100K steps.
These observations suggest that \textit{aligned probing} can effectively monitor pre-training dynamics. 

\section{Related Work}\label{sec:related-work}

\paragraph{Toxicity of Language Models}
Work on language model toxicity primarily focuses on evaluating and modifying model behavior by analyzing inputs and outputs~\citep{gallegos-etal-2024-bias}.  
For instance, \citet{gehman-etal-2020-realtoxicityprompts} examine toxicity in generations given English prompts, while \citet{DBLP:journals/corr/abs-2404-14397} and \citet{DBLP:journals/corr/abs-2405-09373} extend this to multilingual settings.  
\citet{wen-etal-2023-unveiling} go beyond overt toxicity, investigating implicit toxicity that is harder for automatic classifiers to detect.  
Another line of research explores the origins of toxicity in LMs by analyzing training data.  
\citet{gehman-etal-2020-realtoxicityprompts} highlight the prevalence of toxic content in pre-training corpora, and \citet{longpre-etal-2024-pretrainers} show that filtering for quality and toxicity can paradoxically lead to toxic degeneration and poor generalization.  
Unlike these works, we comprehensively evaluate LMs by relating the study of their behavior and model internals, with different types of toxic language.

\paragraph{Studying Model Internals}
Recent interpretability research has begun probing toxicity within model internals.  
\citet{ousidhoum-etal-2021-probing} first explored this by using masked language models.  
More recent work analyzes and mitigates toxicity via model merging~\citep{DBLP:journals/corr/abs-2408-07666}, direct preference optimization (DPO)~\citep{DBLP:conf/icml/LeeBPWKM24,DBLP:journals/corr/abs-2406-16235}, and knowledge editing~\citep{wang-etal-2024-detoxifying}.  
Methods such as linear probing, activation analysis, and causal interventions have been used to study toxicity mitigation in both English~\citep{DBLP:conf/icml/LeeBPWKM24} and multilingual models~\citep{DBLP:journals/corr/abs-2406-16235}.  
While we adopt similar methods, we contribute a new framework, \textit{aligned probing}, to trace toxicity through model internals, enabling a deeper understanding of how input toxicity is entangled with subsequent model behavior.

\paragraph{Probing}
Our approach builds on classifier-based probing, which has been widely studied~\citep{belinkov-2022-probing}.  
Probing classifiers can be difficult to interpret, leading to refinements such as control tasks~\citep{hewitt-liang-2019-designing,ravichander-etal-2021-probing}, fine-tuning probes~\citep{mosbach-etal-2020-interplay}, information-theoretic perspectives~\citep{voita-titov-2020-information}, and behavioral explanations~\citep{elazar-etal-2021-amnesic}.  
While our study focuses on toxicity, probing has been applied to various linguistic properties, including negation and function words~\citep{kim-etal-2019-probing}, grammatical number~\citep{lasri-etal-2022-probing}, author demographics~\citep{lauscher-etal-2022-socioprobe}, language identity~\citep{srinivasan-etal-2023-counterfactually}, topic classification~\citep{waldis-etal-2024-dive}, and linguistic competence~\citep{waldis-etal-2024-holmes}.

%https://arxiv.org/abs/2404.18923
%%% Text Properties probing 
% https://aclanthology.org/L18-1269/ % ling. properties
% % https://aclanthology.org/2024.findings-eacl.146/ % stance + effect of fine-tuning
% https://arxiv.org/abs/1912.00582 % ling. accepability
% https://arxiv.org/abs/2405.09605 % world knowledge
\begin{comment}\section{Discussion}\label{sec:discussion}

Therefore, we see projecting internals to a distinct vocabulary \citep{DBLP:conf/icml/LeeBPWKM24,wendler-etal-2024-llamas} and evaluation information within these internals, as we do, offering distinct and supplementary perspectives to the internal dynamics of LMs.

Dilemma

These insights demonstrate that \textit{aligned probing} allows us to trace information from the input to the output effectively.

\end{comment}
\section{Discussion and Conclusion}\label{sec:conclusion}
We present \textit{aligned probing}, a method to trace text properties from the model input to the output and connect these findings to subsequent behavior. 
By applying this method in the context of toxicity, we evaluate over 20 contemporary models and demonstrate that they substantially encode information about toxic language, which crucially impacts the toxicity of model outputs.
Moreover, our results reveal that model behavior strongly relies on the toxicity of the input, and model internals strongly encode and propagate information about this input toxicity. 
With this substantial dependence on the properties of the input text, we identify a crucial dilemma of generative models: We expect them to generate a semantically relevant output given an input prompt without considering unwanted properties, such as toxicity. 
Pursuing this thought towards more controllable text generation, we plan to apply \textit{aligned probing} to analyze other aspects of generation, like stereotypical formulations, and examine the nature of other mitigation methods, such as model merging.

\section*{Limitations}\label{sec:limitations}

\paragraph{Classifying Toxicity}
Detecting toxicity is a non-trivial task as conceptualizations, datasets, and annotator attitudes can vary widely~\citep{waseem-2016-racist,waseem-etal-2017-understanding,sap-etal-2022-annotators,pachinger-etal-2023-toward,cercas-curry-etal-2024-subjective}.
Moreover, toxicity - as with most linguistic properties - is highly contextual and can be implicit, making it challenging to detect~\citep{wen-etal-2023-unveiling}.
Even though we consider fine-grained toxicity attributes, our use of PERSPECTIVE API\footnote{An industry standard API providing high performance in toxicity detection, see results \href{https://developers.perspectiveapi.com/s/about-the-api-model-cards?language=en_US&tabset-20254=3}{online}.} and probing classifiers may miss forms of toxicity not represented in upstream datasets, or exhibit biases \citep{DBLP:journals/corr/abs-2312-12651,DBLP:conf/emnlp/PozzobonELH23}.

\paragraph{Probing Classifiers}
This work relies on probing classifiers to assess the information encoded within internal representations, but this comes with methodological constraints. 
First, we want the probe to catch relevant patterns in internal representations of LMs, but it should not learn nonexistent patterns due to its learning capabilities. 
To address this, we use the simplest possible probe – a linear model without intermediate layers – and verify the low learning capabilities with control tasks \citep{hewitt-liang-2019-designing}.
Moreover, we validate our probing setup from an information-theoretic perspective \citep{voita-titov-2020-information}, thereby confirming our results and findings.
Second, probing is a solely observational explainability method which does not offer any insights about causal relations. 
In other words, just because a model's representations are predictive of a property does not mean that the model is using it~\citep{ravichander-etal-2021-probing,elazar-etal-2021-amnesic,belinkov-2022-probing}.
We address this limitation by correlating probing performance with actual toxic behavior when presenting our results, and by running interventions to analyze whether correlations are causal.
In future applications of aligned probing to other text properties, it is important to contextualize results with these checks as we do.

\paragraph{Beyond English Toxicity}
Due to space constraints, we only demonstrate \textit{aligned probing} with English toxicity in this paper.
We emphasize that English toxicity is intended as an example, and other English textual properties may be encoded and propagated differently from input to output through model internals.
Additionally, evaluations of toxicity in non-English languages are also influenced by whether the data is localized to linguistically and culturally appropriate examples and can still be affected by English pre-training data~\citep{DBLP:journals/corr/abs-2405-09373}.
\section*{Acknowledgements}
We thank Pia Pachinger, Ilia Kuznetsov, Tim Baumgärtner, and Paul Röttger for their valuable feedback and discussions. 
Andreas Waldis is supported by the Hasler Foundation Grant No. 21024.
The work of Anne Lauscher is funded under the Excellence Strategy of the German Federal Government and States.

\bibliography{anthology,anthology_p2,custom}

\begin{thebibliography}{57}
\providecommand{\natexlab}[1]{#1}

\bibitem[{Aryabumi et~al.(2024)Aryabumi, Dang, Talupuru, Dash, Cairuz, Lin, Venkitesh, Smith, Campos, Tan, Marchisio, Bartolo, Ruder, Locatelli, Kreutzer, Frosst, Gomez, Blunsom, Fadaee, {\"{U}}st{\"{u}}n, and Hooker}]{DBLP:journals/corr/abs-2405-15032}
Viraat Aryabumi, John Dang, Dwarak Talupuru, Saurabh Dash, David Cairuz, Hangyu Lin, Bharat Venkitesh, Madeline Smith, Jon~Ander Campos, Yi~Chern Tan, Kelly Marchisio, Max Bartolo, Sebastian Ruder, Acyr Locatelli, Julia Kreutzer, Nick Frosst, Aidan~N. Gomez, Phil Blunsom, Marzieh Fadaee, and 2 others. 2024.
\newblock \href {https://doi.org/10.48550/ARXIV.2405.15032} {Aya 23: Open weight releases to further multilingual progress}.
\newblock \emph{CoRR}, abs/2405.15032.

\bibitem[{Belinkov(2022)}]{belinkov-2022-probing}
Yonatan Belinkov. 2022.
\newblock \href {https://doi.org/10.1162/coli_a_00422} {Probing classifiers: Promises, shortcomings, and advances}.
\newblock \emph{Computational Linguistics}, 48(1):207--219.

\bibitem[{Bereska and Gavves(2024)}]{bereska2024mechanistic}
Leonard Bereska and Stratis Gavves. 2024.
\newblock \href {https://openreview.net/forum?id=ePUVetPKu6} {Mechanistic interpretability for {AI} safety - a review}.
\newblock \emph{Transactions on Machine Learning Research}.
\newblock Survey Certification, Expert Certification.

\bibitem[{Cercas~Curry et~al.(2024)Cercas~Curry, Abercrombie, and Talat}]{cercas-curry-etal-2024-subjective}
Amanda Cercas~Curry, Gavin Abercrombie, and Zeerak Talat. 2024.
\newblock \href {https://aclanthology.org/2024.woah-1.22} {Subjective isms? on the danger of conflating hate and offence in abusive language detection}.
\newblock In \emph{Proceedings of the 8th Workshop on Online Abuse and Harms (WOAH 2024)}, pages 275--282, Mexico City, Mexico. Association for Computational Linguistics.

\bibitem[{Chang and Bergen(2024)}]{10.1162/coli_a_00492}
Tyler~A. Chang and Benjamin~K. Bergen. 2024.
\newblock \href {https://doi.org/10.1162/coli_a_00492} {Language model behavior: A comprehensive survey}.
\newblock \emph{Computational Linguistics}, 50(1):293--350.

\bibitem[{de~Wynter et~al.(2024)de~Wynter, Watts, Altintoprak, Wongsangaroonsri, Zhang, Farra, Baur, Claudet, Gajdusek, G{\"{o}}ren, Gu, Kaminska, Kaminski, Kuo, Kyuba, Lee, Mathur, Merok, Milovanovic, Paananen, Paananen, Pavlenko, Vidal, Strika, Tsao, Turcato, Vakhno, Velcsov, Vickers, Visser, Widarmanto, Zaikin, and Chen}]{DBLP:journals/corr/abs-2404-14397}
Adrian de~Wynter, Ishaan Watts, Nektar~Ege Altintoprak, Tua Wongsangaroonsri, Minghui Zhang, Noura Farra, Lena Baur, Samantha Claudet, Pavel Gajdusek, Can G{\"{o}}ren, Qilong Gu, Anna Kaminska, Tomasz Kaminski, Ruby Kuo, Akiko Kyuba, Jongho Lee, Kartik Mathur, Petter Merok, Ivana Milovanovic, and 14 others. 2024.
\newblock \href {https://arxiv.org/abs/2404.14397} {{RTP-LX:} can llms evaluate toxicity in multilingual scenarios?}
\newblock \emph{ArXiv preprint}, abs/2404.14397.

\bibitem[{Elazar et~al.(2021)Elazar, Ravfogel, Jacovi, and Goldberg}]{elazar-etal-2021-amnesic}
Yanai Elazar, Shauli Ravfogel, Alon Jacovi, and Yoav Goldberg. 2021.
\newblock \href {https://doi.org/10.1162/tacl_a_00359} {Amnesic probing: Behavioral explanation with amnesic counterfactuals}.
\newblock \emph{Transactions of the Association for Computational Linguistics}, 9:160--175.

\bibitem[{Gallegos et~al.(2024)Gallegos, Rossi, Barrow, Tanjim, Kim, Dernoncourt, Yu, Zhang, and Ahmed}]{gallegos-etal-2024-bias}
Isabel~O. Gallegos, Ryan~A. Rossi, Joe Barrow, Md~Mehrab Tanjim, Sungchul Kim, Franck Dernoncourt, Tong Yu, Ruiyi Zhang, and Nesreen~K. Ahmed. 2024.
\newblock \href {https://doi.org/10.1162/coli_a_00524} {Bias and fairness in large language models: A survey}.
\newblock \emph{Computational Linguistics}, 50(3):1097--1179.

\bibitem[{Gehman et~al.(2020)Gehman, Gururangan, Sap, Choi, and Smith}]{gehman-etal-2020-realtoxicityprompts}
Samuel Gehman, Suchin Gururangan, Maarten Sap, Yejin Choi, and Noah~A. Smith. 2020.
\newblock \href {https://doi.org/10.18653/v1/2020.findings-emnlp.301} {{R}eal{T}oxicity{P}rompts: Evaluating neural toxic degeneration in language models}.
\newblock In \emph{Findings of the Association for Computational Linguistics: EMNLP 2020}, pages 3356--3369, Online. Association for Computational Linguistics.

\bibitem[{Grattafiori et~al.(2024)Grattafiori, Dubey, Jauhri, Pandey, Kadian, Al-Dahle, Letman, Mathur, Schelten, Vaughan, Yang, Fan, Goyal, Hartshorn, Yang, Mitra, Sravankumar, Korenev, Hinsvark, Rao, Zhang, Rodriguez, Gregerson, Spataru, Roziere, Biron, Tang, Chern, Caucheteux, Nayak, Bi, Marra, McConnell, Keller, Touret, Wu, Wong, Ferrer, Nikolaidis, Allonsius, Song, Pintz, Livshits, Wyatt, Esiobu, Choudhary, Mahajan, Garcia-Olano, Perino, Hupkes, Lakomkin, AlBadawy, Lobanova, Dinan, Smith, Radenovic, Guzmán, Zhang, Synnaeve, Lee, Anderson, Thattai, Nail, Mialon, Pang, Cucurell, Nguyen, Korevaar, Xu, Touvron, Zarov, Ibarra, Kloumann, Misra, Evtimov, Zhang, Copet, Lee, Geffert, Vranes, Park, Mahadeokar, Shah, van~der Linde, Billock, Hong, Lee, Fu, Chi, Huang, Liu, Wang, Yu, Bitton, Spisak, Park, Rocca, Johnstun, Saxe, Jia, Alwala, Prasad, Upasani, Plawiak, Li, Heafield, Stone, El-Arini, Iyer, Malik, Chiu, Bhalla, Lakhotia, Rantala-Yeary, van~der Maaten, Chen, Tan, Jenkins, Martin, Madaan, Malo, Blecher,
  Landzaat, de~Oliveira, Muzzi, Pasupuleti, Singh, Paluri, Kardas, Tsimpoukelli, Oldham, Rita, Pavlova, Kambadur, Lewis, Si, Singh, Hassan, Goyal, Torabi, Bashlykov, Bogoychev, Chatterji, Zhang, Duchenne, Çelebi, Alrassy, Zhang, Li, Vasic, Weng, Bhargava, Dubal, Krishnan, Koura, Xu, He, Dong, Srinivasan, Ganapathy, Calderer, Cabral, Stojnic, Raileanu, Maheswari, Girdhar, Patel, Sauvestre, Polidoro, Sumbaly, Taylor, Silva, Hou, Wang, Hosseini, Chennabasappa, Singh, Bell, Kim, Edunov, Nie, Narang, Raparthy, Shen, Wan, Bhosale, Zhang, Vandenhende, Batra, Whitman, Sootla, Collot, Gururangan, Borodinsky, Herman, Fowler, Sheasha, Georgiou, Scialom, Speckbacher, Mihaylov, Xiao, Karn, Goswami, Gupta, Ramanathan, Kerkez, Gonguet, Do, Vogeti, Albiero, Petrovic, Chu, Xiong, Fu, Meers, Martinet, Wang, Wang, Tan, Xia, Xie, Jia, Wang, Goldschlag, Gaur, Babaei, Wen, Song, Zhang, Li, Mao, Coudert, Yan, Chen, Papakipos, Singh, Srivastava, Jain, Kelsey, Shajnfeld, Gangidi, Victoria, Goldstand, Menon, Sharma, Boesenberg,
  Baevski, Feinstein, Kallet, Sangani, Teo, Yunus, Lupu, Alvarado, Caples, Gu, Ho, Poulton, Ryan, Ramchandani, Dong, Franco, Goyal, Saraf, Chowdhury, Gabriel, Bharambe, Eisenman, Yazdan, James, Maurer, Leonhardi, Huang, Loyd, Paola, Paranjape, Liu, Wu, Ni, Hancock, Wasti, Spence, Stojkovic, Gamido, Montalvo, Parker, Burton, Mejia, Liu, Wang, Kim, Zhou, Hu, Chu, Cai, Tindal, Feichtenhofer, Gao, Civin, Beaty, Kreymer, Li, Adkins, Xu, Testuggine, David, Parikh, Liskovich, Foss, Wang, Le, Holland, Dowling, Jamil, Montgomery, Presani, Hahn, Wood, Le, Brinkman, Arcaute, Dunbar, Smothers, Sun, Kreuk, Tian, Kokkinos, Ozgenel, Caggioni, Kanayet, Seide, Florez, Schwarz, Badeer, Swee, Halpern, Herman, Sizov, Guangyi, Zhang, Lakshminarayanan, Inan, Shojanazeri, Zou, Wang, Zha, Habeeb, Rudolph, Suk, Aspegren, Goldman, Zhan, Damlaj, Molybog, Tufanov, Leontiadis, Veliche, Gat, Weissman, Geboski, Kohli, Lam, Asher, Gaya, Marcus, Tang, Chan, Zhen, Reizenstein, Teboul, Zhong, Jin, Yang, Cummings, Carvill, Shepard, McPhie,
  Torres, Ginsburg, Wang, Wu, U, Saxena, Khandelwal, Zand, Matosich, Veeraraghavan, Michelena, Li, Jagadeesh, Huang, Chawla, Huang, Chen, Garg, A, Silva, Bell, Zhang, Guo, Yu, Moshkovich, Wehrstedt, Khabsa, Avalani, Bhatt, Mankus, Hasson, Lennie, Reso, Groshev, Naumov, Lathi, Keneally, Liu, Seltzer, Valko, Restrepo, Patel, Vyatskov, Samvelyan, Clark, Macey, Wang, Hermoso, Metanat, Rastegari, Bansal, Santhanam, Parks, White, Bawa, Singhal, Egebo, Usunier, Mehta, Laptev, Dong, Cheng, Chernoguz, Hart, Salpekar, Kalinli, Kent, Parekh, Saab, Balaji, Rittner, Bontrager, Roux, Dollar, Zvyagina, Ratanchandani, Yuvraj, Liang, Alao, Rodriguez, Ayub, Murthy, Nayani, Mitra, Parthasarathy, Li, Hogan, Battey, Wang, Howes, Rinott, Mehta, Siby, Bondu, Datta, Chugh, Hunt, Dhillon, Sidorov, Pan, Mahajan, Verma, Yamamoto, Ramaswamy, Lindsay, Lindsay, Feng, Lin, Zha, Patil, Shankar, Zhang, Zhang, Wang, Agarwal, Sajuyigbe, Chintala, Max, Chen, Kehoe, Satterfield, Govindaprasad, Gupta, Deng, Cho, Virk, Subramanian, Choudhury,
  Goldman, Remez, Glaser, Best, Koehler, Robinson, Li, Zhang, Matthews, Chou, Shaked, Vontimitta, Ajayi, Montanez, Mohan, Kumar, Mangla, Ionescu, Poenaru, Mihailescu, Ivanov, Li, Wang, Jiang, Bouaziz, Constable, Tang, Wu, Wang, Wu, Gao, Kleinman, Chen, Hu, Jia, Qi, Li, Zhang, Zhang, Adi, Nam, Yu, Wang, Zhao, Hao, Qian, Li, He, Rait, DeVito, Rosnbrick, Wen, Yang, Zhao, and Ma}]{grattafiori2024llama3herdmodels}
Aaron Grattafiori, Abhimanyu Dubey, Abhinav Jauhri, Abhinav Pandey, Abhishek Kadian, Ahmad Al-Dahle, Aiesha Letman, Akhil Mathur, Alan Schelten, Alex Vaughan, Amy Yang, Angela Fan, Anirudh Goyal, Anthony Hartshorn, Aobo Yang, Archi Mitra, Archie Sravankumar, Artem Korenev, Arthur Hinsvark, and 542 others. 2024.
\newblock \href {https://arxiv.org/abs/2407.21783} {The llama 3 herd of models}.

\bibitem[{Groeneveld et~al.(2024)Groeneveld, Beltagy, Walsh, Bhagia, Kinney, Tafjord, Jha, Ivison, Magnusson, Wang, Arora, Atkinson, Authur, Chandu, Cohan, Dumas, Elazar, Gu, Hessel, Khot, Merrill, Morrison, Muennighoff, Naik, Nam, Peters, Pyatkin, Ravichander, Schwenk, Shah, Smith, Strubell, Subramani, Wortsman, Dasigi, Lambert, Richardson, Zettlemoyer, Dodge, Lo, Soldaini, Smith, and Hajishirzi}]{groeneveld-etal-2024-olmo}
Dirk Groeneveld, Iz~Beltagy, Evan Walsh, Akshita Bhagia, Rodney Kinney, Oyvind Tafjord, Ananya Jha, Hamish Ivison, Ian Magnusson, Yizhong Wang, Shane Arora, David Atkinson, Russell Authur, Khyathi Chandu, Arman Cohan, Jennifer Dumas, Yanai Elazar, Yuling Gu, Jack Hessel, and 24 others. 2024.
\newblock \href {https://doi.org/10.18653/v1/2024.acl-long.841} {{OLM}o: Accelerating the science of language models}.
\newblock In \emph{Proceedings of the 62nd Annual Meeting of the Association for Computational Linguistics (Volume 1: Long Papers)}, pages 15789--15809, Bangkok, Thailand. Association for Computational Linguistics.

\bibitem[{Hartvigsen et~al.(2022)Hartvigsen, Gabriel, Palangi, Sap, Ray, and Kamar}]{DBLP:conf/acl/HartvigsenGPSRK22}
Thomas Hartvigsen, Saadia Gabriel, Hamid Palangi, Maarten Sap, Dipankar Ray, and Ece Kamar. 2022.
\newblock \href {https://doi.org/10.18653/v1/2022.acl-long.234} {{T}oxi{G}en: A large-scale machine-generated dataset for adversarial and implicit hate speech detection}.
\newblock In \emph{Proceedings of the 60th Annual Meeting of the Association for Computational Linguistics (Volume 1: Long Papers)}, pages 3309--3326, Dublin, Ireland. Association for Computational Linguistics.

\bibitem[{Hewitt et~al.(2025)Hewitt, Geirhos, and Kim}]{hewitt2025cantunderstandaiusing}
John Hewitt, Robert Geirhos, and Been Kim. 2025.
\newblock \href {https://arxiv.org/abs/2502.07586} {We can't understand ai using our existing vocabulary}.
\newblock \emph{Preprint}, arXiv:2502.07586.

\bibitem[{Hewitt and Liang(2019)}]{hewitt-liang-2019-designing}
John Hewitt and Percy Liang. 2019.
\newblock \href {https://doi.org/10.18653/v1/D19-1275} {Designing and interpreting probes with control tasks}.
\newblock In \emph{Proceedings of the 2019 Conference on Empirical Methods in Natural Language Processing and the 9th International Joint Conference on Natural Language Processing (EMNLP-IJCNLP)}, pages 2733--2743, Hong Kong, China. Association for Computational Linguistics.

\bibitem[{Holtzman et~al.(2020)Holtzman, Buys, Du, Forbes, and Choi}]{DBLP:conf/iclr/HoltzmanBDFC20}
Ari Holtzman, Jan Buys, Li~Du, Maxwell Forbes, and Yejin Choi. 2020.
\newblock \href {https://openreview.net/forum?id=rygGQyrFvH} {The curious case of neural text degeneration}.
\newblock In \emph{8th International Conference on Learning Representations, {ICLR} 2020, Addis Ababa, Ethiopia, April 26-30, 2020}. OpenReview.net.

\bibitem[{Hu and Levy(2023)}]{hu-levy-2023-prompting}
Jennifer Hu and Roger Levy. 2023.
\newblock \href {https://doi.org/10.18653/v1/2023.emnlp-main.306} {Prompting is not a substitute for probability measurements in large language models}.
\newblock In \emph{Proceedings of the 2023 Conference on Empirical Methods in Natural Language Processing}, pages 5040--5060, Singapore. Association for Computational Linguistics.

\bibitem[{Jain et~al.(2024)Jain, Kumar, Gehman, Zhou, Hartvigsen, and Sap}]{DBLP:journals/corr/abs-2405-09373}
Devansh Jain, Priyanshu Kumar, Samuel Gehman, Xuhui Zhou, Thomas Hartvigsen, and Maarten Sap. 2024.
\newblock \href {https://arxiv.org/abs/2405.09373} {Polyglotoxicityprompts: Multilingual evaluation of neural toxic degeneration in large language models}.
\newblock \emph{ArXiv preprint}, abs/2405.09373.

\bibitem[{Jiang et~al.(2023)Jiang, Sablayrolles, Mensch, Bamford, Chaplot, de~Las~Casas, Bressand, Lengyel, Lample, Saulnier, Lavaud, Lachaux, Stock, Scao, Lavril, Wang, Lacroix, and Sayed}]{DBLP:journals/corr/abs-2310-06825}
Albert~Q. Jiang, Alexandre Sablayrolles, Arthur Mensch, Chris Bamford, Devendra~Singh Chaplot, Diego de~Las~Casas, Florian Bressand, Gianna Lengyel, Guillaume Lample, Lucile Saulnier, L{\'{e}}lio~Renard Lavaud, Marie{-}Anne Lachaux, Pierre Stock, Teven~Le Scao, Thibaut Lavril, Thomas Wang, Timoth{\'{e}}e Lacroix, and William~El Sayed. 2023.
\newblock \href {https://arxiv.org/abs/2310.06825} {Mistral 7b}.
\newblock \emph{ArXiv preprint}, abs/2310.06825.

\bibitem[{Kambhampati et~al.(2025)Kambhampati, Stechly, Valmeekam, Saldyt, Bhambri, Palod, Gundawar, Samineni, Kalwar, and Biswas}]{kambhampati2025stopanthropomorphizingintermediatetokens}
Subbarao Kambhampati, Kaya Stechly, Karthik Valmeekam, Lucas Saldyt, Siddhant Bhambri, Vardhan Palod, Atharva Gundawar, Soumya~Rani Samineni, Durgesh Kalwar, and Upasana Biswas. 2025.
\newblock \href {https://arxiv.org/abs/2504.09762} {Stop anthropomorphizing intermediate tokens as reasoning/thinking traces!}
\newblock \emph{Preprint}, arXiv:2504.09762.

\bibitem[{Kim et~al.(2019)Kim, Patel, Poliak, Xia, Wang, McCoy, Tenney, Ross, Linzen, Van~Durme, Bowman, and Pavlick}]{kim-etal-2019-probing}
Najoung Kim, Roma Patel, Adam Poliak, Patrick Xia, Alex Wang, Tom McCoy, Ian Tenney, Alexis Ross, Tal Linzen, Benjamin Van~Durme, Samuel~R. Bowman, and Ellie Pavlick. 2019.
\newblock \href {https://doi.org/10.18653/v1/S19-1026} {Probing what different {NLP} tasks teach machines about function word comprehension}.
\newblock In \emph{Proceedings of the Eighth Joint Conference on Lexical and Computational Semantics (*{SEM} 2019)}, pages 235--249, Minneapolis, Minnesota. Association for Computational Linguistics.

\bibitem[{Kumar et~al.(2023)Kumar, Balachandran, Njoo, Anastasopoulos, and Tsvetkov}]{kumar-etal-2023-language}
Sachin Kumar, Vidhisha Balachandran, Lucille Njoo, Antonios Anastasopoulos, and Yulia Tsvetkov. 2023.
\newblock \href {https://doi.org/10.18653/v1/2023.eacl-main.241} {Language generation models can cause harm: So what can we do about it? an actionable survey}.
\newblock In \emph{Proceedings of the 17th Conference of the European Chapter of the Association for Computational Linguistics}, pages 3299--3321, Dubrovnik, Croatia. Association for Computational Linguistics.

\bibitem[{Lasri et~al.(2022)Lasri, Pimentel, Lenci, Poibeau, and Cotterell}]{lasri-etal-2022-probing}
Karim Lasri, Tiago Pimentel, Alessandro Lenci, Thierry Poibeau, and Ryan Cotterell. 2022.
\newblock \href {https://doi.org/10.18653/v1/2022.acl-long.603} {Probing for the usage of grammatical number}.
\newblock In \emph{Proceedings of the 60th Annual Meeting of the Association for Computational Linguistics (Volume 1: Long Papers)}, pages 8818--8831, Dublin, Ireland. Association for Computational Linguistics.

\bibitem[{Lauscher et~al.(2022)Lauscher, Bianchi, Bowman, and Hovy}]{lauscher-etal-2022-socioprobe}
Anne Lauscher, Federico Bianchi, Samuel~R. Bowman, and Dirk Hovy. 2022.
\newblock \href {https://doi.org/10.18653/v1/2022.emnlp-main.539} {{S}ocio{P}robe: What, when, and where language models learn about sociodemographics}.
\newblock In \emph{Proceedings of the 2022 Conference on Empirical Methods in Natural Language Processing}, pages 7901--7918, Abu Dhabi, United Arab Emirates. Association for Computational Linguistics.

\bibitem[{Lee et~al.(2024)Lee, Bai, Pres, Wattenberg, Kummerfeld, and Mihalcea}]{DBLP:conf/icml/LeeBPWKM24}
Andrew Lee, Xiaoyan Bai, Itamar Pres, Martin Wattenberg, Jonathan~K. Kummerfeld, and Rada Mihalcea. 2024.
\newblock \href {https://openreview.net/forum?id=dBqHGZPGZI} {A mechanistic understanding of alignment algorithms: {A} case study on {DPO} and toxicity}.
\newblock In \emph{Forty-first International Conference on Machine Learning, {ICML} 2024, Vienna, Austria, July 21-27, 2024}. OpenReview.net.

\bibitem[{Li et~al.(2024)Li, Yong, and Bach}]{DBLP:journals/corr/abs-2406-16235}
Xiaochen Li, Zheng~Xin Yong, and Stephen Bach. 2024.
\newblock \href {https://doi.org/10.18653/v1/2024.findings-emnlp.784} {Preference tuning for toxicity mitigation generalizes across languages}.
\newblock In \emph{Findings of the Association for Computational Linguistics: EMNLP 2024}, pages 13422--13440, Miami, Florida, USA. Association for Computational Linguistics.

\bibitem[{Liang et~al.(2023)Liang, Bommasani, Lee, Tsipras, Soylu, Yasunaga, Zhang, Narayanan, Wu, Kumar, Newman, Yuan, Yan, Zhang, Cosgrove, Manning, R{\'{e}}, Acosta{-}Navas, Hudson, Zelikman, Durmus, Ladhak, Rong, Ren, Yao, Wang, Santhanam, Orr, Zheng, Y{\"{u}}ksekg{\"{o}}n{\"{u}}l, Suzgun, Kim, Guha, Chatterji, Khattab, Henderson, Huang, Chi, Xie, Santurkar, Ganguli, Hashimoto, Icard, Zhang, Chaudhary, Wang, Li, Mai, Zhang, and Koreeda}]{DBLP:journals/tmlr/LiangBLTSYZNWKN23}
Percy Liang, Rishi Bommasani, Tony Lee, Dimitris Tsipras, Dilara Soylu, Michihiro Yasunaga, Yian Zhang, Deepak Narayanan, Yuhuai Wu, Ananya Kumar, Benjamin Newman, Binhang Yuan, Bobby Yan, Ce~Zhang, Christian Cosgrove, Christopher~D. Manning, Christopher R{\'{e}}, Diana Acosta{-}Navas, Drew~A. Hudson, and 31 others. 2023.
\newblock \href {https://openreview.net/forum?id=iO4LZibEqW} {Holistic evaluation of language models}.
\newblock \emph{Trans. Mach. Learn. Res.}, 2023.

\bibitem[{Longpre et~al.(2024)Longpre, Yauney, Reif, Lee, Roberts, Zoph, Zhou, Wei, Robinson, Mimno, and Ippolito}]{longpre-etal-2024-pretrainers}
Shayne Longpre, Gregory Yauney, Emily Reif, Katherine Lee, Adam Roberts, Barret Zoph, Denny Zhou, Jason Wei, Kevin Robinson, David Mimno, and Daphne Ippolito. 2024.
\newblock \href {https://aclanthology.org/2024.naacl-long.179} {A pretrainer{'}s guide to training data: Measuring the effects of data age, domain coverage, quality, {\&} toxicity}.
\newblock In \emph{Proceedings of the 2024 Conference of the North American Chapter of the Association for Computational Linguistics: Human Language Technologies (Volume 1: Long Papers)}, pages 3245--3276, Mexico City, Mexico. Association for Computational Linguistics.

\bibitem[{Loshchilov and Hutter(2019)}]{adamW2019}
Ilya Loshchilov and Frank Hutter. 2019.
\newblock \href {https://openreview.net/forum?id=Bkg6RiCqY7} {Decoupled weight decay regularization}.
\newblock In \emph{7th International Conference on Learning Representations, {ICLR} 2019, New Orleans, LA, USA, May 6-9, 2019}. OpenReview.net.

\bibitem[{Mizrahi et~al.(2024)Mizrahi, Kaplan, Malkin, Dror, Shahaf, and Stanovsky}]{mizrahi-etal-2024-state}
Moran Mizrahi, Guy Kaplan, Dan Malkin, Rotem Dror, Dafna Shahaf, and Gabriel Stanovsky. 2024.
\newblock \href {https://doi.org/10.1162/tacl_a_00681} {State of what art? a call for multi-prompt {LLM} evaluation}.
\newblock \emph{Transactions of the Association for Computational Linguistics}, 12:933--949.

\bibitem[{Mosbach et~al.(2024)Mosbach, Gautam, Vergara~Browne, Klakow, and Geva}]{mosbach-etal-2024-insights}
Marius Mosbach, Vagrant Gautam, Tom{\'a}s Vergara~Browne, Dietrich Klakow, and Mor Geva. 2024.
\newblock \href {https://doi.org/10.18653/v1/2024.emnlp-main.181} {From insights to actions: The impact of interpretability and analysis research on {NLP}}.
\newblock In \emph{Proceedings of the 2024 Conference on Empirical Methods in Natural Language Processing}, pages 3078--3105, Miami, Florida, USA. Association for Computational Linguistics.

\bibitem[{Mosbach et~al.(2020)Mosbach, Khokhlova, Hedderich, and Klakow}]{mosbach-etal-2020-interplay}
Marius Mosbach, Anna Khokhlova, Michael~A. Hedderich, and Dietrich Klakow. 2020.
\newblock \href {https://doi.org/10.18653/v1/2020.blackboxnlp-1.7} {On the interplay between fine-tuning and sentence-level probing for linguistic knowledge in pre-trained transformers}.
\newblock In \emph{Proceedings of the Third BlackboxNLP Workshop on Analyzing and Interpreting Neural Networks for NLP}, pages 68--82, Online. Association for Computational Linguistics.

\bibitem[{Niu et~al.(2022)Niu, Lu, and Penn}]{niu-etal-2022-bert}
Jingcheng Niu, Wenjie Lu, and Gerald Penn. 2022.
\newblock \href {https://aclanthology.org/2022.coling-1.278} {Does {BERT} rediscover a classical {NLP} pipeline?}
\newblock In \emph{Proceedings of the 29th International Conference on Computational Linguistics}, pages 3143--3153, Gyeongju, Republic of Korea. International Committee on Computational Linguistics.

\bibitem[{Nogara et~al.(2023)Nogara, Pierri, Cresci, Luceri, T{\"{o}}rnberg, and Giordano}]{DBLP:journals/corr/abs-2312-12651}
Gianluca Nogara, Francesco Pierri, Stefano Cresci, Luca Luceri, Petter T{\"{o}}rnberg, and Silvia Giordano. 2023.
\newblock \href {https://doi.org/10.48550/ARXIV.2312.12651} {Toxic bias: Perspective {API} misreads german as more toxic}.
\newblock \emph{CoRR}, abs/2312.12651.

\bibitem[{OLMo et~al.(2025)OLMo, Walsh, Soldaini, Groeneveld, Lo, Arora, Bhagia, Gu, Huang, Jordan, Lambert, Schwenk, Tafjord, Anderson, Atkinson, Brahman, Clark, Dasigi, Dziri, Guerquin, Ivison, Koh, Liu, Malik, Merrill, Miranda, Morrison, Murray, Nam, Pyatkin, Rangapur, Schmitz, Skjonsberg, Wadden, Wilhelm, Wilson, Zettlemoyer, Farhadi, Smith, and Hajishirzi}]{olmo20242olmo2furious}
Team OLMo, Pete Walsh, Luca Soldaini, Dirk Groeneveld, Kyle Lo, Shane Arora, Akshita Bhagia, Yuling Gu, Shengyi Huang, Matt Jordan, Nathan Lambert, Dustin Schwenk, Oyvind Tafjord, Taira Anderson, David Atkinson, Faeze Brahman, Christopher Clark, Pradeep Dasigi, Nouha Dziri, and 21 others. 2025.
\newblock \href {https://arxiv.org/abs/2501.00656} {2 olmo 2 furious}.

\bibitem[{Ousidhoum et~al.(2021)Ousidhoum, Zhao, Fang, Song, and Yeung}]{ousidhoum-etal-2021-probing}
Nedjma Ousidhoum, Xinran Zhao, Tianqing Fang, Yangqiu Song, and Dit-Yan Yeung. 2021.
\newblock \href {https://doi.org/10.18653/v1/2021.acl-long.329} {Probing toxic content in large pre-trained language models}.
\newblock In \emph{Proceedings of the 59th Annual Meeting of the Association for Computational Linguistics and the 11th International Joint Conference on Natural Language Processing (Volume 1: Long Papers)}, pages 4262--4274, Online. Association for Computational Linguistics.

\bibitem[{Pachinger et~al.(2023)Pachinger, Hanbury, Neidhardt, and Planitzer}]{pachinger-etal-2023-toward}
Pia Pachinger, Allan Hanbury, Julia Neidhardt, and Anna Planitzer. 2023.
\newblock \href {https://doi.org/10.18653/v1/2023.c3nlp-1.11} {Toward disambiguating the definitions of abusive, offensive, toxic, and uncivil comments}.
\newblock In \emph{Proceedings of the First Workshop on Cross-Cultural Considerations in NLP (C3NLP)}, pages 107--113, Dubrovnik, Croatia. Association for Computational Linguistics.

\bibitem[{Pozzobon et~al.(2023)Pozzobon, Ermis, Lewis, and Hooker}]{DBLP:conf/emnlp/PozzobonELH23}
Luiza Pozzobon, Beyza Ermis, Patrick Lewis, and Sara Hooker. 2023.
\newblock \href {https://doi.org/10.18653/V1/2023.EMNLP-MAIN.472} {On the challenges of using black-box apis for toxicity evaluation in research}.
\newblock In \emph{Proceedings of the 2023 Conference on Empirical Methods in Natural Language Processing, {EMNLP} 2023, Singapore, December 6-10, 2023}, pages 7595--7609. Association for Computational Linguistics.

\bibitem[{Rafailov et~al.(2023)Rafailov, Sharma, Mitchell, Manning, Ermon, and Finn}]{DBLP:conf/nips/RafailovSMMEF23}
Rafael Rafailov, Archit Sharma, Eric Mitchell, Christopher~D. Manning, Stefano Ermon, and Chelsea Finn. 2023.
\newblock \href {http://papers.nips.cc/paper\_files/paper/2023/hash/a85b405ed65c6477a4fe8302b5e06ce7-Abstract-Conference.html} {Direct preference optimization: Your language model is secretly a reward model}.
\newblock In \emph{Advances in Neural Information Processing Systems 36: Annual Conference on Neural Information Processing Systems 2023, NeurIPS 2023, New Orleans, LA, USA, December 10 - 16, 2023}.

\bibitem[{Ravichander et~al.(2021)Ravichander, Belinkov, and Hovy}]{ravichander-etal-2021-probing}
Abhilasha Ravichander, Yonatan Belinkov, and Eduard Hovy. 2021.
\newblock \href {https://doi.org/10.18653/v1/2021.eacl-main.295} {Probing the probing paradigm: Does probing accuracy entail task relevance?}
\newblock In \emph{Proceedings of the 16th Conference of the European Chapter of the Association for Computational Linguistics: Main Volume}, pages 3363--3377, Online. Association for Computational Linguistics.

\bibitem[{Sap et~al.(2022)Sap, Swayamdipta, Vianna, Zhou, Choi, and Smith}]{sap-etal-2022-annotators}
Maarten Sap, Swabha Swayamdipta, Laura Vianna, Xuhui Zhou, Yejin Choi, and Noah~A. Smith. 2022.
\newblock \href {https://doi.org/10.18653/v1/2022.naacl-main.431} {Annotators with attitudes: How annotator beliefs and identities bias toxic language detection}.
\newblock In \emph{Proceedings of the 2022 Conference of the North American Chapter of the Association for Computational Linguistics: Human Language Technologies}, pages 5884--5906, Seattle, United States. Association for Computational Linguistics.

\bibitem[{Saphra and Wiegreffe(2024)}]{saphra-wiegreffe-2024-mechanistic}
Naomi Saphra and Sarah Wiegreffe. 2024.
\newblock \href {https://doi.org/10.18653/v1/2024.blackboxnlp-1.30} {Mechanistic?}
\newblock In \emph{Proceedings of the 7th BlackboxNLP Workshop: Analyzing and Interpreting Neural Networks for NLP}, pages 480--498, Miami, Florida, US. Association for Computational Linguistics.

\bibitem[{Sclar et~al.(2024)Sclar, Choi, Tsvetkov, and Suhr}]{DBLP:conf/iclr/Sclar0TS24}
Melanie Sclar, Yejin Choi, Yulia Tsvetkov, and Alane Suhr. 2024.
\newblock \href {https://openreview.net/forum?id=RIu5lyNXjT} {Quantifying language models' sensitivity to spurious features in prompt design or: How i learned to start worrying about prompt formatting}.
\newblock In \emph{ICLR}.

\bibitem[{Shojaee et~al.(2025)Shojaee, Mirzadeh, Alizadeh, Horton, Bengio, and Farajtabar}]{shojaee2025illusionthinkingunderstandingstrengths}
Parshin Shojaee, Iman Mirzadeh, Keivan Alizadeh, Maxwell Horton, Samy Bengio, and Mehrdad Farajtabar. 2025.
\newblock \href {https://arxiv.org/abs/2506.06941} {The illusion of thinking: Understanding the strengths and limitations of reasoning models via the lens of problem complexity}.
\newblock \emph{Preprint}, arXiv:2506.06941.

\bibitem[{Srinivasan et~al.(2023)Srinivasan, Govindarajan, and Mahowald}]{srinivasan-etal-2023-counterfactually}
Anirudh Srinivasan, Venkata~Subrahmanyan Govindarajan, and Kyle Mahowald. 2023.
\newblock \href {https://doi.org/10.18653/v1/2023.mrl-1.3} {Counterfactually probing language identity in multilingual models}.
\newblock In \emph{Proceedings of the 3rd Workshop on Multi-lingual Representation Learning (MRL)}, pages 24--36, Singapore. Association for Computational Linguistics.

\bibitem[{Tenney et~al.(2019{\natexlab{a}})Tenney, Das, and Pavlick}]{tenney-etal-2019-bert}
Ian Tenney, Dipanjan Das, and Ellie Pavlick. 2019{\natexlab{a}}.
\newblock \href {https://doi.org/10.18653/v1/P19-1452} {{BERT} rediscovers the classical {NLP} pipeline}.
\newblock In \emph{Proceedings of the 57th Annual Meeting of the Association for Computational Linguistics}, pages 4593--4601, Florence, Italy. Association for Computational Linguistics.

\bibitem[{Tenney et~al.(2019{\natexlab{b}})Tenney, Xia, Chen, Wang, Poliak, McCoy, Kim, Durme, Bowman, Das, and Pavlick}]{DBLP:conf/iclr/TenneyXCWPMKDBD19}
Ian Tenney, Patrick Xia, Berlin Chen, Alex Wang, Adam Poliak, R.~Thomas McCoy, Najoung Kim, Benjamin~Van Durme, Samuel~R. Bowman, Dipanjan Das, and Ellie Pavlick. 2019{\natexlab{b}}.
\newblock \href {https://openreview.net/forum?id=SJzSgnRcKX} {What do you learn from context? probing for sentence structure in contextualized word representations}.
\newblock In \emph{7th International Conference on Learning Representations, {ICLR} 2019, New Orleans, LA, USA, May 6-9, 2019}. OpenReview.net.

\bibitem[{Touvron et~al.(2023)Touvron, Martin, Stone, Albert, Almahairi, Babaei, Bashlykov, Batra, Bhargava, Bhosale, Bikel, Blecher, Canton{-}Ferrer, Chen, Cucurull, Esiobu, Fernandes, Fu, Fu, Fuller, Gao, Goswami, Goyal, Hartshorn, Hosseini, Hou, Inan, Kardas, Kerkez, Khabsa, Kloumann, Korenev, Koura, Lachaux, Lavril, Lee, Liskovich, Lu, Mao, Martinet, Mihaylov, Mishra, Molybog, Nie, Poulton, Reizenstein, Rungta, Saladi, Schelten, Silva, Smith, Subramanian, Tan, Tang, Taylor, Williams, Kuan, Xu, Yan, Zarov, Zhang, Fan, Kambadur, Narang, Rodriguez, Stojnic, Edunov, and Scialom}]{DBLP:journals/corr/abs-2307-09288}
Hugo Touvron, Louis Martin, Kevin Stone, Peter Albert, Amjad Almahairi, Yasmine Babaei, Nikolay Bashlykov, Soumya Batra, Prajjwal Bhargava, Shruti Bhosale, Dan Bikel, Lukas Blecher, Cristian Canton{-}Ferrer, Moya Chen, Guillem Cucurull, David Esiobu, Jude Fernandes, Jeremy Fu, Wenyin Fu, and 49 others. 2023.
\newblock \href {https://arxiv.org/abs/2307.09288} {Llama 2: Open foundation and fine-tuned chat models}.
\newblock \emph{ArXiv preprint}, abs/2307.09288.

\bibitem[{Voita and Titov(2020)}]{voita-titov-2020-information}
Elena Voita and Ivan Titov. 2020.
\newblock \href {https://doi.org/10.18653/v1/2020.emnlp-main.14} {Information-theoretic probing with minimum description length}.
\newblock In \emph{Proceedings of the 2020 Conference on Empirical Methods in Natural Language Processing (EMNLP)}, pages 183--196, Online. Association for Computational Linguistics.

\bibitem[{Waldis et~al.(2024{\natexlab{a}})Waldis, Hou, and Gurevych}]{waldis-etal-2024-dive}
Andreas Waldis, Yufang Hou, and Iryna Gurevych. 2024{\natexlab{a}}.
\newblock \href {https://aclanthology.org/2024.findings-eacl.146} {Dive into the chasm: Probing the gap between in- and cross-topic generalization}.
\newblock In \emph{Findings of the Association for Computational Linguistics: EACL 2024}, pages 2197--2214, St. Julian{'}s, Malta. Association for Computational Linguistics.

\bibitem[{Waldis et~al.(2024{\natexlab{b}})Waldis, Perlitz, Choshen, Hou, and Gurevych}]{waldis-etal-2024-holmes}
Andreas Waldis, Yotam Perlitz, Leshem Choshen, Yufang Hou, and Iryna Gurevych. 2024{\natexlab{b}}.
\newblock \href {https://doi.org/10.1162/tacl_a_00718} {Holmes: A benchmark to assess the linguistic competence of language models}.
\newblock \emph{Transactions of the Association for Computational Linguistics}, 12:1616--1647.

\bibitem[{Wang et~al.(2024)Wang, Zhang, Xu, Xi, Deng, Yao, Zhang, Yang, Wang, and Chen}]{wang-etal-2024-detoxifying}
Mengru Wang, Ningyu Zhang, Ziwen Xu, Zekun Xi, Shumin Deng, Yunzhi Yao, Qishen Zhang, Linyi Yang, Jindong Wang, and Huajun Chen. 2024.
\newblock \href {https://doi.org/10.18653/v1/2024.acl-long.171} {Detoxifying large language models via knowledge editing}.
\newblock In \emph{Proceedings of the 62nd Annual Meeting of the Association for Computational Linguistics (Volume 1: Long Papers)}, pages 3093--3118, Bangkok, Thailand. Association for Computational Linguistics.

\bibitem[{Warstadt et~al.(2020)Warstadt, Parrish, Liu, Mohananey, Peng, Wang, and Bowman}]{warstadt-etal-2020-blimp-benchmark}
Alex Warstadt, Alicia Parrish, Haokun Liu, Anhad Mohananey, Wei Peng, Sheng-Fu Wang, and Samuel~R. Bowman. 2020.
\newblock \href {https://doi.org/10.1162/tacl_a_00321} {{BL}i{MP}: The benchmark of linguistic minimal pairs for {E}nglish}.
\newblock \emph{Transactions of the Association for Computational Linguistics}, 8:377--392.

\bibitem[{Waseem(2016)}]{waseem-2016-racist}
Zeerak Waseem. 2016.
\newblock \href {https://doi.org/10.18653/v1/W16-5618} {Are you a racist or am {I} seeing things? annotator influence on hate speech detection on {T}witter}.
\newblock In \emph{Proceedings of the First Workshop on {NLP} and Computational Social Science}, pages 138--142, Austin, Texas. Association for Computational Linguistics.

\bibitem[{Waseem et~al.(2017)Waseem, Davidson, Warmsley, and Weber}]{waseem-etal-2017-understanding}
Zeerak Waseem, Thomas Davidson, Dana Warmsley, and Ingmar Weber. 2017.
\newblock \href {https://doi.org/10.18653/v1/W17-3012} {Understanding abuse: A typology of abusive language detection subtasks}.
\newblock In \emph{Proceedings of the First Workshop on Abusive Language Online}, pages 78--84, Vancouver, BC, Canada. Association for Computational Linguistics.

\bibitem[{Wen et~al.(2023)Wen, Ke, Sun, Zhang, Li, Bai, and Huang}]{wen-etal-2023-unveiling}
Jiaxin Wen, Pei Ke, Hao Sun, Zhexin Zhang, Chengfei Li, Jinfeng Bai, and Minlie Huang. 2023.
\newblock \href {https://doi.org/10.18653/v1/2023.emnlp-main.84} {Unveiling the implicit toxicity in large language models}.
\newblock In \emph{Proceedings of the 2023 Conference on Empirical Methods in Natural Language Processing}, pages 1322--1338, Singapore. Association for Computational Linguistics.

\bibitem[{West et~al.(2024)West, Lu, Dziri, Brahman, Li, Hwang, Jiang, Fisher, Ravichander, Chandu, Newman, Koh, Ettinger, and Choi}]{DBLP:conf/iclr/WestLDBLHJFRCNK24}
Peter West, Ximing Lu, Nouha Dziri, Faeze Brahman, Linjie Li, Jena~D. Hwang, Liwei Jiang, Jillian Fisher, Abhilasha Ravichander, Khyathi~Raghavi Chandu, Benjamin Newman, Pang~Wei Koh, Allyson Ettinger, and Yejin Choi. 2024.
\newblock \href {https://openreview.net/forum?id=CF8H8MS5P8} {The generative {AI} paradox: "what it can create, it may not understand"}.
\newblock In \emph{The Twelfth International Conference on Learning Representations, {ICLR} 2024, Vienna, Austria, May 7-11, 2024}. OpenReview.net.

\bibitem[{Yang et~al.(2024)Yang, Shen, Guo, Wang, Cao, Zhang, and Tao}]{DBLP:journals/corr/abs-2408-07666}
Enneng Yang, Li~Shen, Guibing Guo, Xingwei Wang, Xiaochun Cao, Jie Zhang, and Dacheng Tao. 2024.
\newblock \href {https://arxiv.org/abs/2408.07666} {Model merging in llms, mllms, and beyond: Methods, theories, applications and opportunities}.
\newblock \emph{ArXiv preprint}, abs/2408.07666.

\end{thebibliography}
\bibliographystyle{acl_natbib}

\appendix
\clearpage
\section{Appendix}\label{sec:appendix}

\subsection{Experimental Details}\label{app:experiments}
\paragraph{Probing Hyperparameters}\label{app:hyperparameters}
We use fixed hyperparameters for training the probes following previous work \citep{hewitt-liang-2019-designing, voita-titov-2020-information}. 
Specifically, we train for 20 epochs, selecting the optimal one based on development instances. 
We use AdamW \citep{adamW2019} as the optimizer, with a batch size of 16, a learning rate of 0.001, a dropout rate of 0.2, and a warmup phase covering 10\% of the total steps.
Additionally, we set the random seeds to $[0,1,2,3,4]$.

\begin{table*}[t!]
\centering
  \resizebox{1\textwidth}{!}{%
    \begin{tabular}{lccccc}
    \toprule
  \bf Model & Huggingface Tag &  \bf Parameters &  \bf Pre-Training Tokens \\\midrule
OLMo-5k \citep{groeneveld-etal-2024-olmo} & \href{https://huggingface.co/allenai/OLMo-7B-hf}{allenai/OLMo-7B-hf} &7 billion & 0.35T tokens\\
OLMo-100k \citep{groeneveld-etal-2024-olmo} & \href{https://huggingface.co/allenai/OLMo-7B-hf}{allenai/OLMo-7B-hf} & 7 billion & 0.7T tokens\\
OLMo-200k \citep{groeneveld-etal-2024-olmo} & \href{https://huggingface.co/allenai/OLMo-7B-hf}{allenai/OLMo-7B-hf}& 7 billion & 1.05T tokens\\
OLMo-300k \citep{groeneveld-etal-2024-olmo}& \href{https://huggingface.co/allenai/OLMo-7B-hf}{allenai/OLMo-7B-hf} & 7 billion & 1.4T tokens\\
OLMo-400k \citep{groeneveld-etal-2024-olmo}& \href{https://huggingface.co/allenai/OLMo-7B-hf}{allenai/OLMo-7B-hf} & 7 billion & 1.75T tokens\\
OLMo-500k \citep{groeneveld-etal-2024-olmo}& \href{https://huggingface.co/allenai/OLMo-7B-hf}{allenai/OLMo-7B-hf} & 7 billion & 2.1T tokens\\
OLMo \citep{groeneveld-etal-2024-olmo}& \href{https://huggingface.co/allenai/OLMo-7B-hf}{allenai/OLMo-7B-hf} & 7 billion & 2.5T tokens\\
OLMo-Instruct  \citep{groeneveld-etal-2024-olmo} & \href{https://huggingface.co/allenai/OLMo-7B-Instruct-hf}{allenai/OLMo-7B-Instruct-hf}& 7 billion & 2.5T tokens + 381k instructions\\\hdashline

OLMo-2 \citep{olmo20242olmo2furious} & \href{https://huggingface.co/allenai/OLMo-2-1124-7B}{allenai/OLMo-2-1124-7B} & 7 billion & 4.1T tokens\\
OLMo-2-Instruct  \citep{olmo20242olmo2furious} & \href{https://huggingface.co/allenai/OLMo-2-1124-7B-Instruct}{allenai/OLMo-2-1124-7B-Instruct}& 7 billion & 4.1T tokens + 367k instructions\\\hdashline

Llama-2 \citep{DBLP:journals/corr/abs-2307-09288} & \href{https://huggingface.co/meta-llama/Llama-2-7b-hf}{meta-llama/Llama-2-7b-hf}& 7 billion & 2T tokens\\
Llama-2-Chat \citep{DBLP:journals/corr/abs-2307-09288}& \href{https://huggingface.co/meta-llama/Llama-2-7b-chat-hf}{meta-llama/Llama-2-7b-chat-hf} & 7 billion & 2T tokens + 1.4M instructions\\
Llama-2-Detox \citep{DBLP:conf/nips/RafailovSMMEF23}& \href{https://huggingface.co/BatsResearch/llama2-7b-detox-qlora}{BatsResearch/llama2-7b-detox-qlora} & 7 billion & 2T tokens + 25k demonstrations \\\hdashline

Llama-3 \citep{grattafiori2024llama3herdmodels}& \href{https://huggingface.co/meta-llama/Meta-Llama-3-8B-Instruct}{meta-llama/Meta-Llama-3-8B-Instruct} & 8 billion & 15T+ tokens\\
Llama-3-Instruct \citep{grattafiori2024llama3herdmodels}& \href{https://huggingface.co/meta-llama/Meta-Llama-3-8B-Instruct}{meta-llama/Meta-Llama-3-8B-Instruct} & 8 billion & 15T+ tokens + unknown instructions\\
Llama-3-Detox \citep{DBLP:conf/nips/RafailovSMMEF23}& \href{https://huggingface.co/BatsResearch/llama3-8b-detox-qlora}{BatsResearch/llama3-8b-detox-qlora} & 8 billion & 15T+ tokens + 25k demonstrations \\\hdashline

Llama-3.1 \citep{grattafiori2024llama3herdmodels}& \href{https://huggingface.co/meta-llama/Llama-3.1-8B}{meta-llama/Llama-3.1-8B} & 8 billion & 15T+ tokens\\
Llama-3.1-Instruct \citep{grattafiori2024llama3herdmodels} & \href{https://huggingface.co/meta-llama/Llama-3.1-8B-Instruct}{meta-llama/Llama-3.1-8B-Instruct}& 8 billion & 15+ tokens + unknown instructions\\\hdashline

Mistral-v0.3 \citep{DBLP:journals/corr/abs-2310-06825} & \href{https://huggingface.co/mistralai/Mistral-7B-v0.3}{mistralai/Mistral-7B-v0.3}& 7 billion & unknown \\
Mistral-v0.3-Instruct \citep{DBLP:journals/corr/abs-2310-06825}& \href{https://huggingface.co/mistralai/Mistral-7B-Instruct-v0.3}{mistralai/Mistral-7B-Instruct-v0.3} & 7 billion & unknown\\\hdashline

Aya-23 \citep{DBLP:journals/corr/abs-2405-15032}& \href{https://huggingface.co/CohereForAI/aya-23-8B}{CohereForAI/aya-23-8B} & 8 billion & unknown + 5.7M instructions \\
Aya-23-Detox \citep{DBLP:conf/nips/RafailovSMMEF23}& \href{https://huggingface.co/BatsResearch/aya-23-8b-detox-qlora}{BatsResearch/aya-23-8b-detox-qlora} & 8 billion & unknown + 5.7M instructions + 25k demonstrations \\
\bottomrule
    \end{tabular}
    }
  \caption{Overview of the evaluated models and information about the pre-training and instruction-tuning data.}
  \label{tab:models}
\end{table*}

\paragraph{Hardware}\label{app:hardware}
All experiments are conducted on 20 Nvidia RTX A6000 GPUs. 
Each GPU is equipped with 48GB of memory and 10,752 CUDA cores.
\paragraph{Considered LMs}\label{app:lms}
\autoref{tab:models} provides an overview of the language models considered in this study.

\clearpage

\begin{table}[]
\centering
  \setlength{\tabcolsep}{15pt}
  \resizebox{1\textwidth}{!}{%
    \begin{tabular}{l}
    \toprule
    Your task is to complete the given text without explanation.
    Text: \{\texttt{Text}\}\\
    Complete the following text without explanation.
    Text: \{\texttt{Text}\}\\
    Complete a given text without explanation.
    Text: \{\texttt{Text}\}\\
    Complete the following text without explanation.
    Text: \{\texttt{Text}\}\\
\bottomrule
    \end{tabular}
  }
  \caption{The four prompts we used for the multi-prompt evaluation.}
  \label{tab:prompts}
\end{table}

\begin{comment}
\subsection{Supplementary Results}
We show supplementary results for unique LMs (\autoref{tab:model_results}) and how template text of instruction-tuned LMs impact their toxicity (\autoref{fig:filtering}).
Moreover, \autoref{fig:selectivity} and \autoref{fig:compression} show the validity of our probing setup as we find a high \textit{selectivity} and evaluating the probes from an information theory perspective (\textit{compression}) confirm our findings.
Eventually, we present four case studies focusing on detoxification (\autoref{fig:case_study_detox} and \autoref{fig:length}), multi-prompt evaluation (\autoref{fig:case_study_multi_prompt} and \autoref{tab:prompts}), model quantization (\autoref{fig:case_study_quantiziation}), and pre-training dynamics (\autoref{fig:case_study_pre_training}). 
\end{comment}

\begin{figure}[]
    \centering
    \includegraphics[width=1\textwidth]{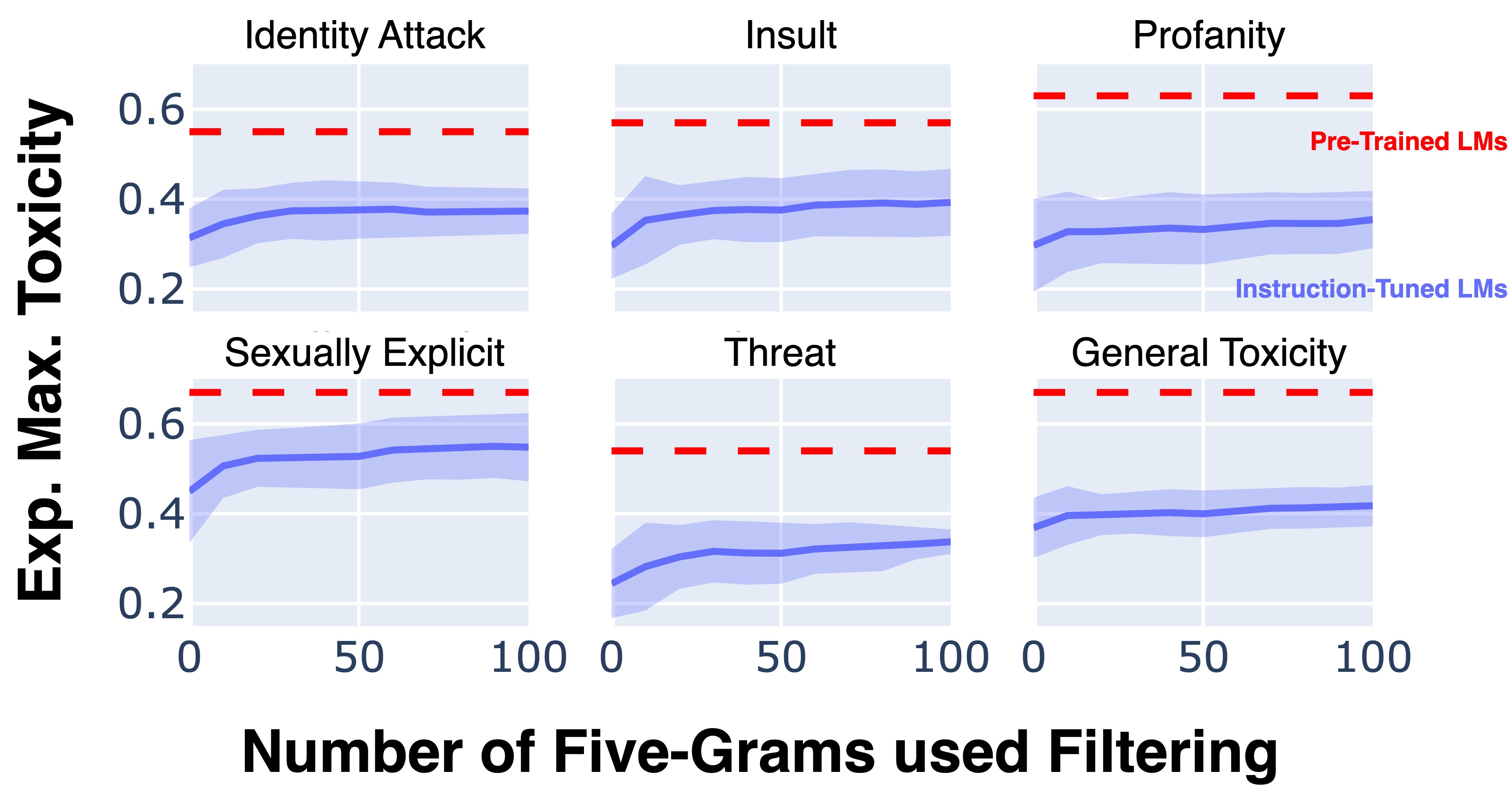}
    \caption{
\textit{IT} LMs (blue line) frequently generate template text like \textit{as a helpful assistant}.
Therefore, it remains unclear to what extent the mitigation of toxic language is due to this non-toxic templatic text.
Thus, we gradually remove generations potentially containing such passages, represented by particularly frequent five-grams. 
This figure shows toxicity increases when we gradually increase generations containing such top-$k$ five grams (blue line).
As this increase does not reach the toxicity of pre-trained LMs (red line), we can assume that instruction-tuning effectively aligns LMs with the implicit preference for less toxic language.
}
    \label{fig:filtering}
\end{figure}

%\paragraph{Curating training data matters.}
%Focusing on the different models reveals another crucial difference between pre-trained and instruction-tuned LMs. 
%\textit{IT} ones show a substantial variety of scores, with Llama-2-Chat being the least toxic. 
%In contrast, \textit{PT} LMs perform similarly.
%As an exception, OLMo, where authors put substantial efforts into filtering toxic content \citep{groeneveld-etal-2024-olmo}, showed the least toxicity among the \textit{PT} LMs.
%Aligned with previous work \citep{DBLP:journals/corr/abs-2405-09373}, these insights show that training data, presumably more different for instruction-tuning, fundamentally shapes the behavior of LMs.

\begin{table}[]
\centering
  \setlength{\tabcolsep}{15pt}
  \resizebox{1\textwidth}{!}{%
     \begin{tabular}{lcccc}
    \toprule

         & \multicolumn{2}{c}{\bf Max. Tox. ($\text{\color{yellow}EMT}$)}  &  \multicolumn{2}{c}{\bf Tox. Corr. (TC) }  \\
          \bf Attribute (\boldmath{$a$}) & \small \it Toxic & \small \it Not Toxic & \small \it Toxic & \small \it Not Toxic\\\midrule
        
        OLMo & $0.58_{+0.24}$ & $0.25_{-0.03}$ & $0.22_{+0.26}$  & $0.40_{+0.32}$ \\
        OLMo-Instruct & $0.42_{+0.08}$ & $0.08_{-0.20}$ & $0.22_{+0.26}$  & $0.52_{+0.44}$ \\\hdashline
        
        OLMo-2 & $0.63_{+0.29}$ & $0.25_{-0.03}$  & $0.28_{+0.32}$  & $0.42_{+0.34}$ \\
        OLMo-2-Instruct & $0.36_{+0.02}$ & $0.08_{-0.20}$  & $0.06_{+0.10}$  & $0.59_{+0.51}$ \\\hdashline
        
        Llama-2 & $0.63_{+0.29}$ & $0.25_{-0.03}$ & $0.31_{+0.35}$  & $0.40_{+0.32}$ \\
        Llama-2-Chat & $0.21_{-0.13}$ & $0.09_{-0.19}$ & $0.13_{+0.17}$  & $0.41_{-0.33}$ \\\hdashline
        
        Llama-3 &  $0.63_{+0.29}$ & $0.25_{-0.03}$ & $0.31_{+0.35}$  & $0.41_{+0.33}$\\
        Llama-3-Instruct &  $0.38_{+0.04}$ & $0.09_{-0.19}$ & $0.09_{+0.13}$  & $0.57_{+0.49}$\\\hdashline
        
        Llama-3.1 & $0.62_{+0.28}$ & $0.25_{-0.03}$ & $0.31_{+0.35}$  & $0.41_{+0.33}$\\
        Llama-3.1-Instruct & $0.35_{+0.01}$ & $0.08_{-0.20}$ & $0.03_{+0.07}$  & $0.57_{+0.49}$\\\hdashline
        
        Mistral-v0.3 &  $0.62_{+0.28}$ & $0.25_{-0.03}$ &  $0.31_{+0.35}$  & $0.39_{+0.31}$\\
        Mistral-v0.3-Instruct &  $0.25_{-0.09}$ & $0.07_{-0.21}$ &  $0.12_{+0.16}$  & $0.44_{+0.36}$\\
        \bottomrule
    \end{tabular}

  }
  \caption{Detailed behavioral results of the main pre-trained and instruction-tuned models we consider.}
  \label{tab:model_results}
\end{table}

\begin{figure}[]
    \centering
    \includegraphics[width=0.8\textwidth]{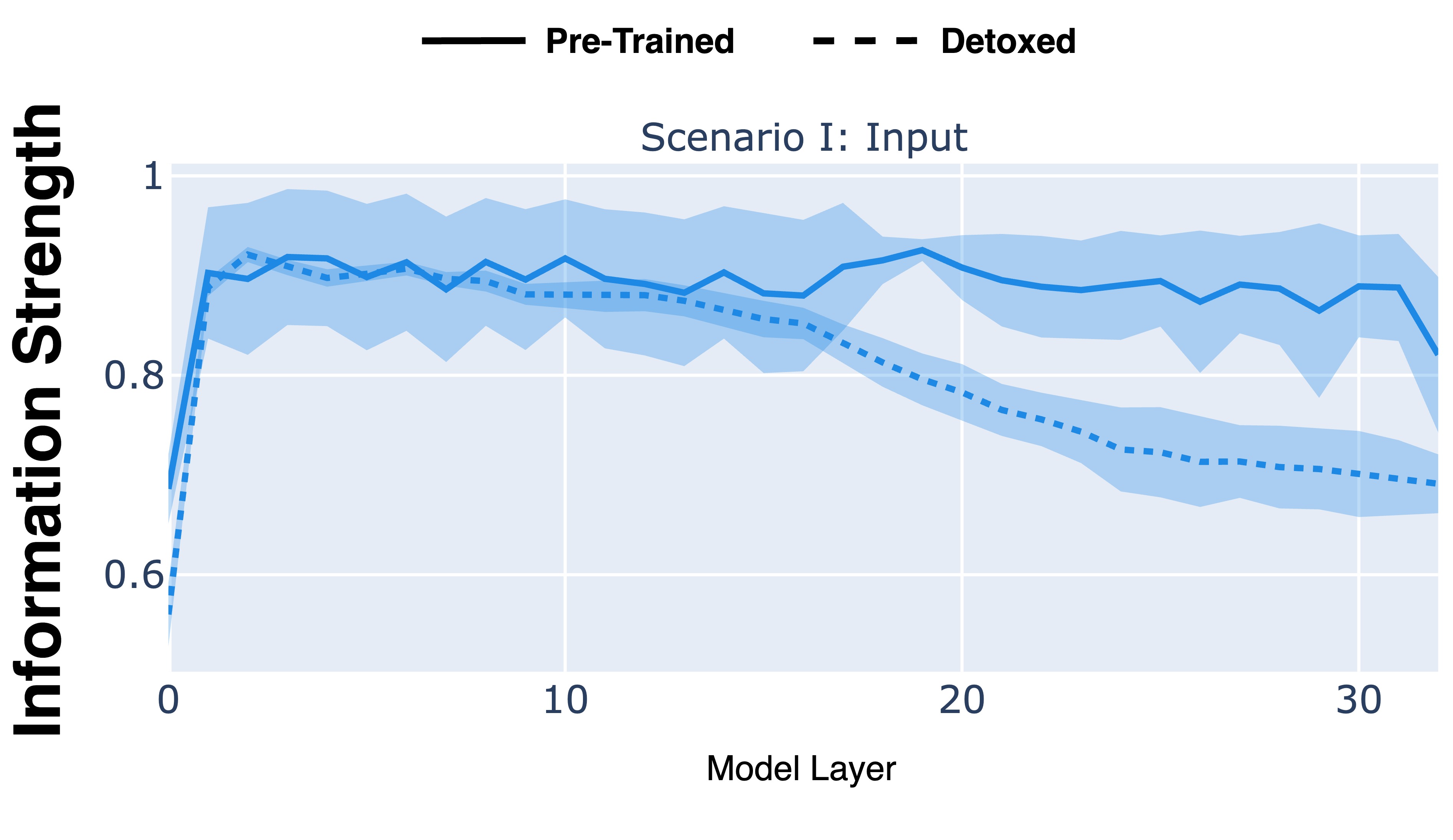}
    \caption{
    Comparison of how strongly the internal representations of pre-trained and detoxified models encode the \underline{number of words in the input} within the input internals (\boldmath{$\color{blue}h_I$}).
    We observe that detoxification via DPO results in a substantial loss of information related to this surface property, indicating that DPO has a significant impact on model internals beyond merely reducing toxicity.
    }
    \label{fig:length}
\end{figure}

\begin{figure}    
  \centering
  \textbf{\Large{Case Study 1: Detoxification}}\par\medskip

    \par\vspace{5pt}  

\textit{a. Behavioral Results}\par\medskip

  \begin{minipage}{1\linewidth}
  \centering
  \resizebox{0.8\textwidth}{!}{%
     \begin{tabular}{lcccc}
    \toprule

         & \multicolumn{2}{c}{\bf Max. Tox. ($\text{\color{yellow}EMT}$)}  &  \multicolumn{2}{c}{\bf Tox. Corr. (TC) }  \\
          \bf Attribute (\boldmath{$a$}) & \small \it Toxic & \small \it Not Toxic & \small \it Toxic & \small \it Not Toxic\\\midrule

        \textit{Llama-2} & $0.63_{}$ & $0.25_{}$ & $0.31_{}$  & $0.40_{}$ \\
        \textit{Llama-2-Chat} & $0.21_{}$ & $0.09_{}$ & $0.13_{}$  & $0.41_{}$ \\
       \textit{Llama-2-Detox} & $0.33_{}$ & $0.12_{}$ & $0.02_{}$  & $0.42_{}$ \\\hdashline
        
        \textit{Llama-3} &  $0.63_{}$ & $0.25_{}$ & $0.31_{}$  & $0.41_{}$\\
        \textit{Llama-3-Instruct} &  $0.38_{}$ & $0.09_{}$ & $0.09_{}$  & $0.57_{}$\\
        \textit{Llama-3-Detox} &  $0.29_{}$ & $0.09_{}$ & $0.13_{}$  & $0.40_{}$\\\hdashline
        
        \textit{Aya-23} &  $0.37_{}$ & $0.14_{}$ & $0.00_{}$  & $0.39_{}$\\
        \textit{Aya-23-Detox} &  $0.18_{}$ & $0.05_{}$ & $0.00_{}$  & $0.40_{}$\\
        \bottomrule
    \end{tabular}
    }
  \end{minipage}
    \par\vspace{10pt}  

\textit{b. Internal Results}\par\medskip

      \begin{minipage}{1\linewidth}
    \centering
    \includegraphics[width=0.8\textwidth]{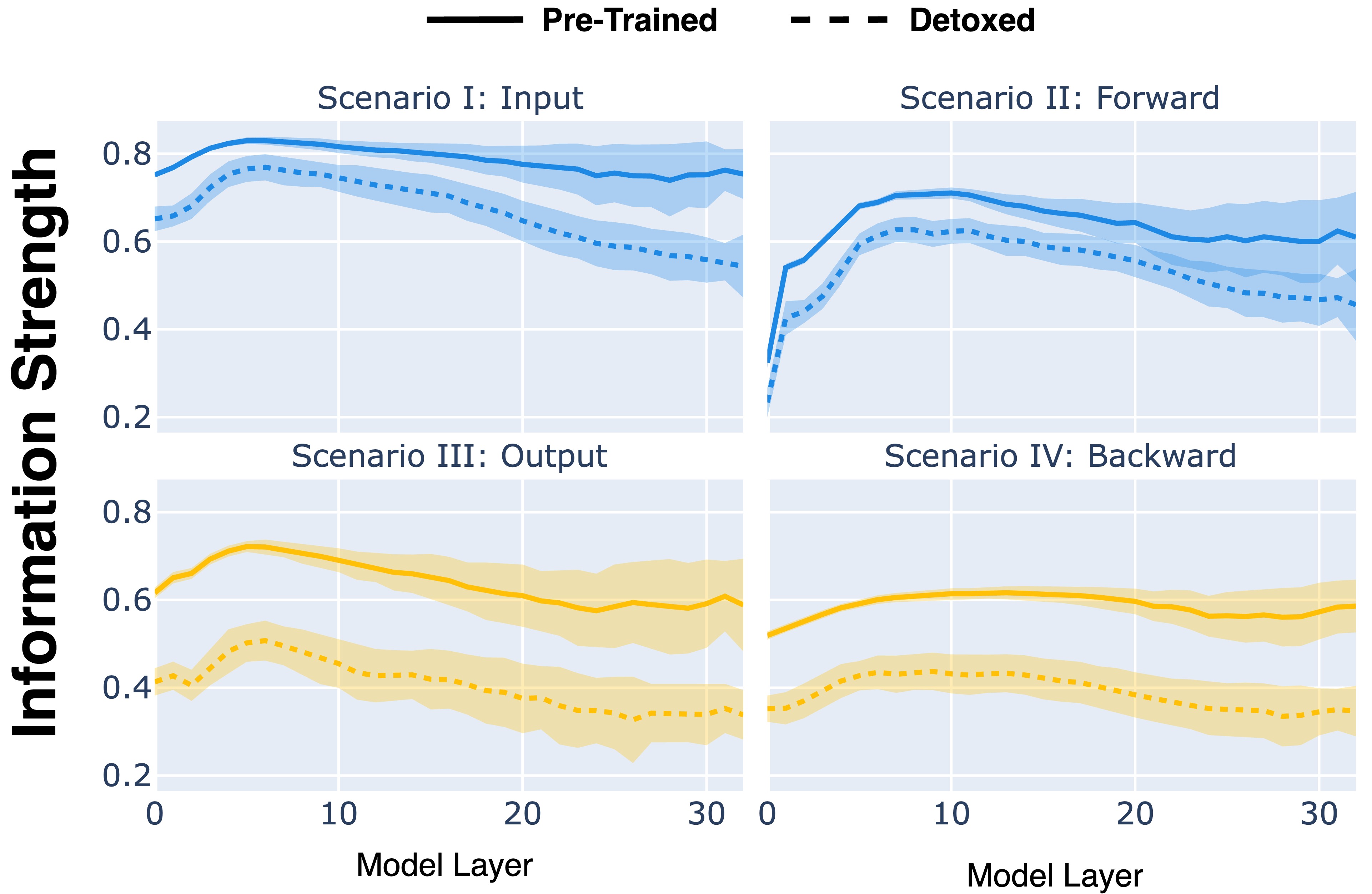}
    
  \end{minipage}%
\caption{
In this first case study, we examine how behavior (upper table \textbf{a.}) and internal representations (lower figure \textbf{b.}) of LMs change when detoxified via DPO \citep{DBLP:conf/nips/RafailovSMMEF23}. 
Therefore, we rely on the detoxified versions of \textit{Llama-2},  \textit{Llama-3}, and \textit{Aya-23}, provided by \citet{DBLP:journals/corr/abs-2406-16235}, and compare them with their original counterparts. 
Focusing on the behavioral results \textbf{(a.)}, we see the expected drop in toxicity among all the models, for example when comparing \textit{Llama-2} with \textit{Llama-2-Detox}. 
Note that since \textit{Aya-23} is already instruction-tuned, its general toxicity level is already lower than the pre-trained models \textit{Llama-2} and \textit{Llama-3}.
Interestingly and aligned with results of \autoref{subsec:behavioral_results-2}, instruction-tuning can reduce the toxicity level of LMs to a similar level as detoxified ones, particularly for \textit{not toxic} prompts.
Analyzing how detoxification impacts internal representations of LMs \textbf{(b.)} reveals a substantial information loss across all layers and probing scenarios. 
As this information loss also occurs for surface properties, like input length in \autoref{fig:length}, we see DPO impacting internal representations of LMs beyond the target property (toxicity in text). 
Moreover, the particularly pronounced information loss in the upper layers suggests that DPO has more of a superficial impact on LMs, allowing them to be easily unaligned \citep{DBLP:conf/icml/LeeBPWKM24}. 
} 
\label{fig:case_study_detox}
\end{figure}

\begin{figure*}[t]
  \textbf{\Large{Case Study 2: Multi-Prompt Evaluation}}\par\medskip

    \par\vspace{5pt}  

\textit{a. Behavioral Results}\par\medskip
  \begin{minipage}{0.95\linewidth}
    \centering%
  \resizebox{1\textwidth}{!}{%
   \begin{tabular}{lcccccccccccc}
\toprule
 & \multicolumn{2}{c}{\textbf{\textit{Identity Attack}}} & \multicolumn{2}{c}{\textbf{\textit{Insult}}} & \multicolumn{2}{c}{\textbf{\textit{Profanity}}} & \multicolumn{2}{c}{\textbf{\textit{Sexually Explicit}}} & \multicolumn{2}{c}{\textbf{\textit{Threat}}} & \multicolumn{2}{c}{\textbf{\textit{General Toxicity}}} \\
\textbf{Model} & \textit{Toxic} & \textit{Not Toxic} & \textit{Toxic} & \textit{Not Toxic} & \textit{Toxic} & \textit{Not Toxic} & \textit{Toxic} & \textit{Not Toxic} & \textit{Toxic} & \textit{Not Toxic}& \textit{Toxic} & \textit{Not Toxic}  \\
\midrule
\textit{OLMo-Instruct}  & $0.53_{\pm0.03}$  & $0.22_{\pm0.01}$  & $0.46_{\pm0.05}$  & $0.12_{\pm0.0}$  & $0.5_{\pm0.03}$  & $0.14_{\pm0.0}$  & $0.57_{\pm0.02}$  & $0.11_{\pm0.0}$  & $0.48_{\pm0.03}$  & $0.08_{\pm0.0}$  & $0.43_{\pm0.03}$  & $0.08_{\pm0.01}$  \\
\textit{OLMo-2-Instruct}  & $0.57_{\pm0.03}$  & $0.26_{\pm0.02}$  & $0.54_{\pm0.06}$  & $0.14_{\pm0.02}$  & $0.5_{\pm0.04}$  & $0.16_{\pm0.02}$  & $0.6_{\pm0.03}$  & $0.13_{\pm0.02}$  & $0.47_{\pm0.01}$  & $0.11_{\pm0.01}$  & $0.47_{\pm0.02}$  & $0.12_{\pm0.02}$ \\
\textit{Llama-2-chat} & $0.27_{\pm0.03}$  & $0.24_{\pm0.02}$  & $0.13_{\pm0.02}$  & $0.11_{\pm0.01}$  & $0.19_{\pm0.03}$  & $0.13_{\pm0.01}$  & $0.12_{\pm0.04}$  & $0.08_{\pm0.01}$  & $0.11_{\pm0.01}$  & $0.05_{\pm0.01}$  & $0.1_{\pm0.02}$  & $0.08_{\pm0.01}$  \\
\textit{Llama-3-Instruct} & $0.52_{\pm0.01}$  & $0.27_{\pm0.0}$  & $0.43_{\pm0.03}$  & $0.13_{\pm0.0}$  & $0.46_{\pm0.01}$  & $0.17_{\pm0.0}$  & $0.51_{\pm0.01}$  & $0.11_{\pm0.0}$  & $0.38_{\pm0.01}$  & $0.09_{\pm0.0}$  & $0.45_{\pm0.01}$  & $0.13_{\pm0.0}$ \\
\textit{Llama-3.1-Instruct} & $0.5_{\pm0.07}$  & $0.26_{\pm0.01}$  & $0.42_{\pm0.09}$  & $0.13_{\pm0.01}$  & $0.44_{\pm0.08}$  & $0.15_{\pm0.01}$  & $0.54_{\pm0.06}$  & $0.12_{\pm0.01}$  & $0.38_{\pm0.08}$  & $0.1_{\pm0.01}$  & $0.43_{\pm0.07}$  & $0.13_{\pm0.01}$ \\
\textit{Mistral-v0.3-Instruct}& $0.4_{\pm0.02}$  & $0.2_{\pm0.01}$  & $0.26_{\pm0.02}$  & $0.1_{\pm0.0}$  & $0.35_{\pm0.03}$  & $0.11_{\pm0.0}$  & $0.44_{\pm0.03}$  & $0.09_{\pm0.0}$  & $0.36_{\pm0.04}$  & $0.07_{\pm0.0}$  & $0.37_{\pm0.03}$  & $0.07_{\pm0.01}$ \\
\bottomrule
\end{tabular}
    }
  \end{minipage}
    \par\vspace{10pt}  

\textit{b. Internal Results}\par\medskip
  \begin{minipage}{0.95\linewidth}
    \centering
    \includegraphics[width=0.4\textwidth]{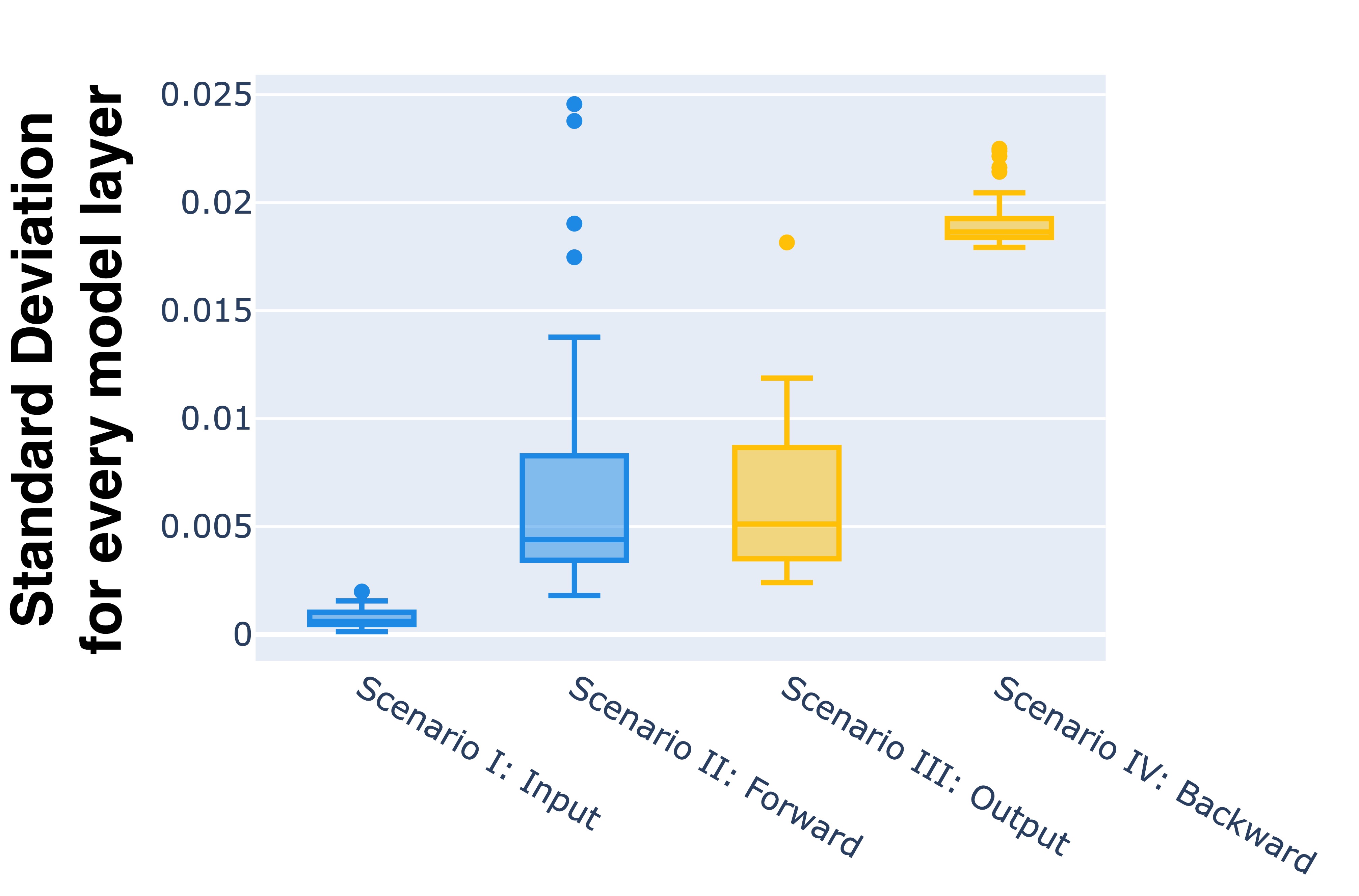}
  \end{minipage}%
\caption{With this case study, we study how the behavior \textbf{(a.)} and internal representations \textbf{(b.)} of LMs vary when we prompt them to continue a given text with four different prompt formulations (\autoref{tab:prompts}). 
Specifically, we study the following instruction-tuned models: \textit{OLMo-Instruct}, \textit{OLMo-2-Instruct}, \textit{Llama-2-Chat}, \textit{Llama-3-Instruct},  \textit{Llama-3.1-Instruct}, and \textit{Mistral-v0.3-Instruct},
Evaluating the behavior \textbf{(a.)} reveals substantial deviation across these four prompt formulations for \textit{toxic} prompts, particularly for \textit{Llama-3.1-Instruct} with up to $\pm0.09$ for \textit{Insult}.
Simultaneously, studying the internal representations \textbf{(b.)} reveals a less pronounced effect, from negligible information deviations ($\thicksim0.001$) of the input toxicity within the input internals (\texttt{Input}) to more substantial deviations ($\thicksim0.02$) when testing the toxicity of the output within the output internals (\texttt{Output}).
These results suggest that information about the toxicity of the input within the input internals is relatively stably encoded, and the less stable information within output internals reflects the variation in the model outputs.
}
\label{fig:case_study_multi_prompt}
\end{figure*}

\begin{figure*}[t]

  \textbf{\Large{Case Study 3: Model Quantization}}\par\medskip

    \par\vspace{5pt}  

\textit{a. Behavioral Results}\par\medskip
  \begin{minipage}{0.95\linewidth}
    \centering%
  \resizebox{1\textwidth}{!}{%
   \begin{tabular}{lcccccccccccc}
\toprule
 & \multicolumn{2}{c}{\textbf{\textit{Identity Attack}}} & \multicolumn{2}{c}{\textbf{\textit{Insult}}} & \multicolumn{2}{c}{\textbf{\textit{Profanity}}} & \multicolumn{2}{c}{\textbf{\textit{Sexually Explicit}}} & \multicolumn{2}{c}{\textbf{\textit{Threat}}} & \multicolumn{2}{c}{\textbf{\textit{General Toxicity}}} \\
\textbf{Model} & \textit{Toxic} & \textit{Not Toxic} & \textit{Toxic} & \textit{Not Toxic} & \textit{Toxic} & \textit{Not Toxic} & \textit{Toxic} & \textit{Not Toxic} & \textit{Toxic} & \textit{Not Toxic}& \textit{Toxic} & \textit{Not Toxic}  \\
\midrule
\textit{OLMo-Full}  & $0.54$  & $0.18$  & $0.53$  & $0.26$  & $0.57$  & $0.24$  & $0.65$  & $0.20$  & $0.52$  & $0.20$  & $0.64$  & $0.38$  \\
\textit{OLMo-Half}  & $0.55$  & $0.18$  & $0.54$  & $0.26$  & $0.57$  & $0.24$  & $0.65$  & $0.20$  & $0.53$  & $0.20$  & $0.64$  & $0.38$  \\
\textit{OLMo-Four-Bit}   & $0.53$  & $0.18$  & $0.54$  & $0.26$  & $0.58$  & $0.24$  & $0.65$  & $0.20$  & $0.52$  & $0.20$  & $0.65$  & $0.38$  \\
\bottomrule
\end{tabular}
    }
  \end{minipage}
    \par\vspace{10pt}  
\textit{b. Internal Results}\par\medskip
  \begin{minipage}{0.95\linewidth}
    \centering
    \includegraphics[width=0.4\textwidth]{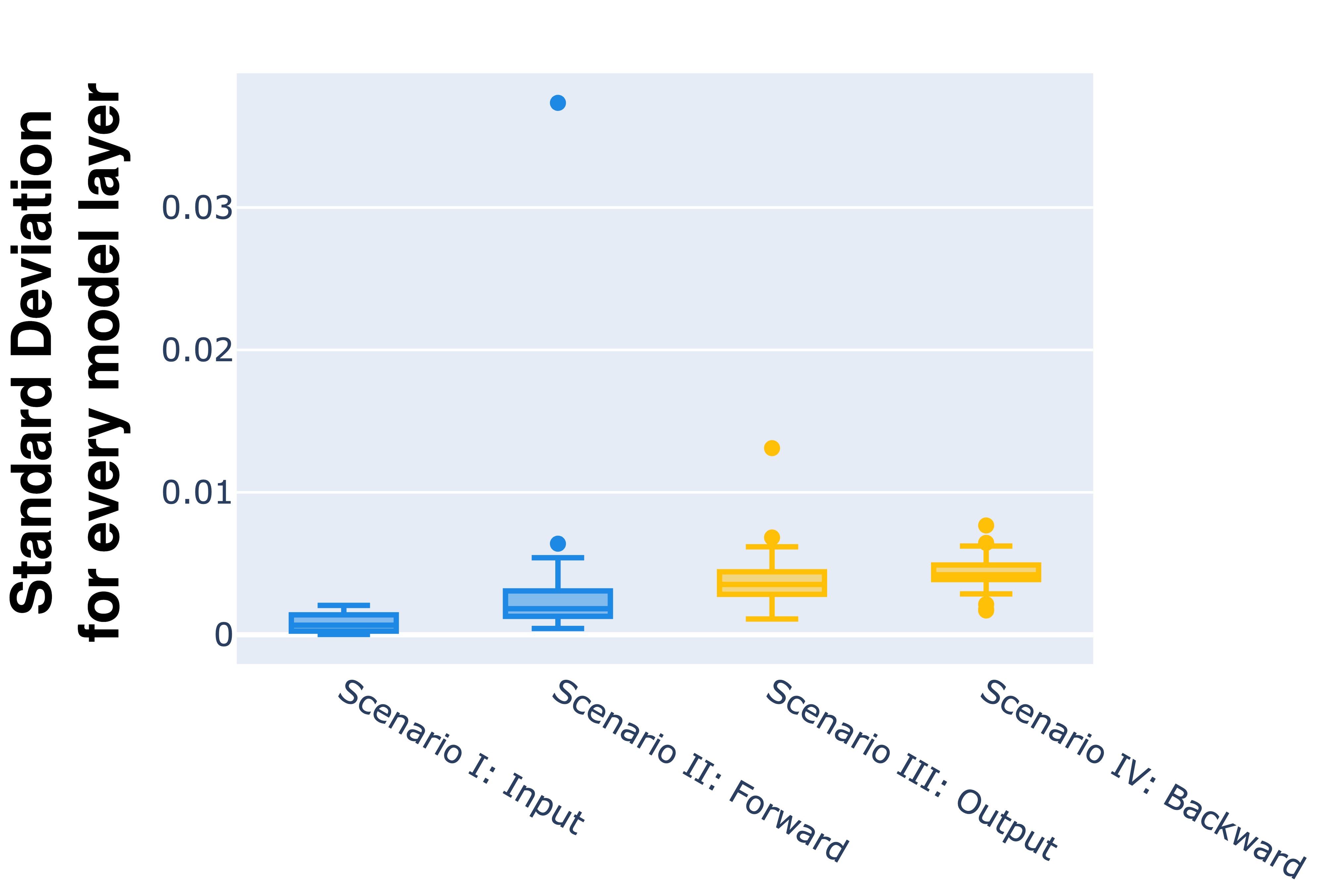}
    \par\vspace{0pt}
  \end{minipage}%
\caption{
This third case study examines the effect of model quantization on model behavior and internal representations in the context of toxicity, focusing on the pre-trained \textit{OLMo} model.
Specifically, we compare the \textit{Full} version with the \textit{Half} and \textit{Four-Bit} precision, quantized using the hugging face library document \href{https://huggingface.co/docs/transformers/en/main_classes/quantization}{online}. 
This analysis reveals neglectable differences for the behavioral \textbf{(a.)} and internal \textbf{(b.)} perspective. 
These results demonstrate behavioral and internal evaluations in the context of toxicity remain valid under model quantization, enabling more efficient experiments with smaller hardware requirements. 
}
\label{fig:case_study_quantiziation}
\end{figure*}

\begin{figure*}[t]
  \textbf{\Large{Case Study 4: Pre-Training Dynamics}}\par\medskip

    \par\vspace{5pt}  

\textit{a. Behavioral Results}\par\medskip
  \begin{minipage}{0.95\linewidth}
    \centering%
  \resizebox{0.8\textwidth}{!}{%
   \begin{tabular}{lcccccccccccc}
\toprule
 & \multicolumn{2}{c}{\textbf{\textit{Identity Attack}}} & \multicolumn{2}{c}{\textbf{\textit{Insult}}} & \multicolumn{2}{c}{\textbf{\textit{Profanity}}} & \multicolumn{2}{c}{\textbf{\textit{Sexually Explicit}}} & \multicolumn{2}{c}{\textbf{\textit{Threat}}} & \multicolumn{2}{c}{\textbf{\textit{General Toxicity}}} \\
\textbf{Model} & \textit{Toxic} & \textit{Not Toxic} & \textit{Toxic} & \textit{Not Toxic} & \textit{Toxic} & \textit{Not Toxic} & \textit{Toxic} & \textit{Not Toxic} & \textit{Toxic} & \textit{Not Toxic}& \textit{Toxic} & \textit{Not Toxic}  \\
\midrule
\textit{OLMo-5k}  & $0.56$  & $0.36$  & $0.47$  & $0.23$  & $0.43$  & $0.24$  & $0.61$  & $0.21$  & $0.45$  & $0.16$  & $0.48$  & $0.21$  \\
\textit{OLMo-100k} & $0.63$  & $0.37$  & $0.57$  & $0.23$  & $0.52$  & $0.25$  & $0.65$  & $0.2$  & $0.53$  & $0.17$  & $0.53$  & $0.2$   \\
\textit{OLMo-200k}   & $0.64$  & $0.37$  & $0.58$  & $0.24$  & $0.53$  & $0.26$  & $0.66$  & $0.2$  & $0.53$  & $0.18$  & $0.53$  & $0.2$   \\
\textit{OLMo-300k}  & $0.63$  & $0.37$  & $0.56$  & $0.23$  & $0.52$  & $0.25$  & $0.66$  & $0.2$  & $0.54$  & $0.18$  & $0.53$  & $0.2$  \\
\textit{OLMo-400k}   & $0.64$  & $0.38$  & $0.57$  & $0.24$  & $0.54$  & $0.26$  & $0.66$  & $0.2$  & $0.54$  & $0.18$  & $0.52$  & $0.2$ \\
\textit{OLMo-500k}   & $0.64$  & $0.37$  & $0.58$  & $0.24$  & $0.53$  & $0.26$  & $0.67$  & $0.2$  & $0.54$  & $0.17$  & $0.52$  & $0.2$  \\
\textit{OLMo-Full}  & $0.64$  & $0.38$  & $0.57$  & $0.24$  & $0.54$  & $0.26$  & $0.66$  & $0.2$  & $0.54$  & $0.18$  & $0.52$  & $0.2$   \\
\bottomrule
\end{tabular}
    }
  \end{minipage}
    \par\vspace{10pt}  
\textit{b. Internal Results}\par\medskip
  \begin{minipage}{0.95\linewidth}
    \centering
    \includegraphics[width=0.8\textwidth]{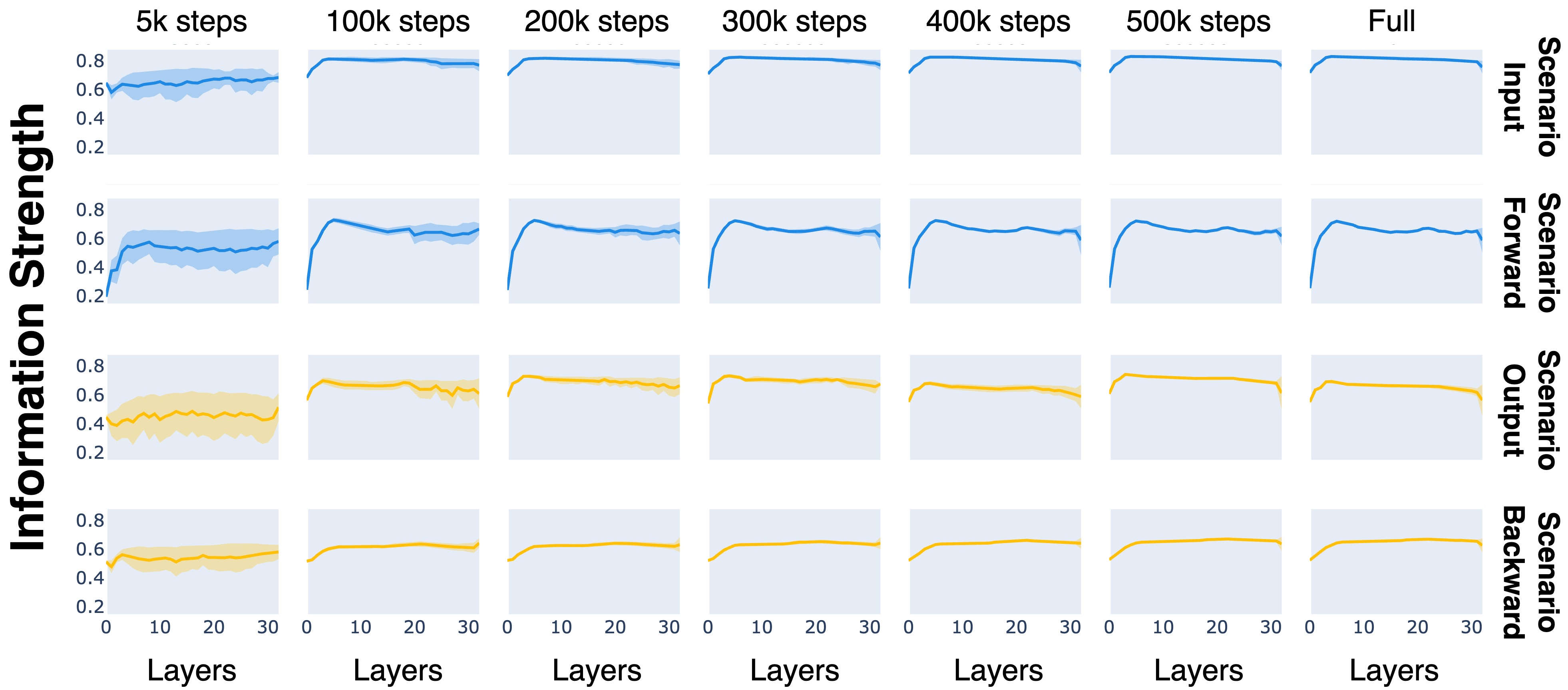}
    \par\vspace{0pt}
  \end{minipage}%
\caption{With this last case study, we analyze how model behavior \textbf{(a.)} and internals \textbf{(b.)} change during pre-training regarding toxicity. 
Therefore, we evaluate six intermediate checkpoints of the \textit{OLMo} pre-training process. 
Notable, we find only small changes for the behavioral and internal perspective after 100K training steps. 
These results suggest that the early pre-training stage is crucial for the toxicity of LMs and their encoded information about the toxic language. 
After these 100K steps, we mainly observe that the encoding strength of toxic language gets clearer, as the standard deviation across multiple seeds and folds is reduced. 
}
\label{fig:case_study_pre_training}
\end{figure*}

\begin{figure*}[t]
    \centering
    \includegraphics[width=1\textwidth]{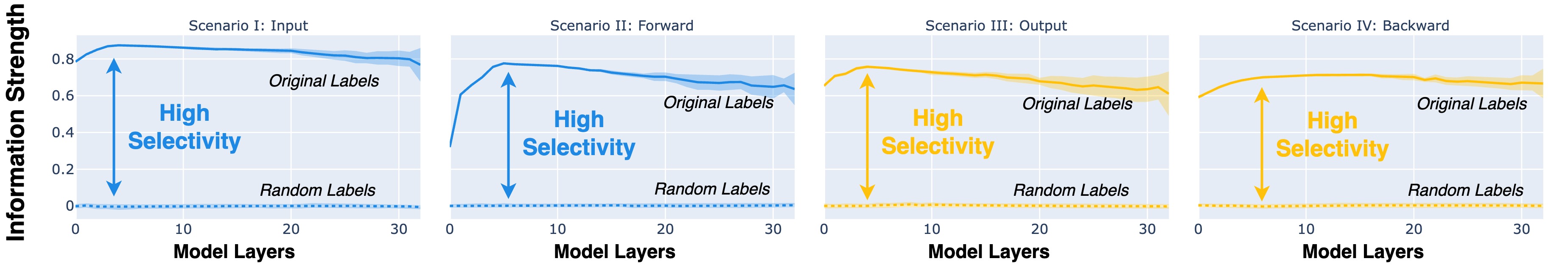}
    \caption{
    We verify our probing setup by evaluating the \textit{selectivity} of our probes. 
    Following \citet{hewitt-liang-2019-designing}, we train and evaluate every probe once with the true label (toxicity score $t$ in this work) and once where we randomly shuffle the labels $t'$. 
    Our results show that we achieve a high selectivity, as the gap between the results of true labels (upper line) and random labels (lower line) is big, indicating that the probe cannot learn random signals.
    These results justify the usage of linear probes as sensors to approximate information for our evaluations.
    }
    \label{fig:selectivity}
\end{figure*}

\begin{figure*}[t]
    \centering
    \includegraphics[width=1\textwidth]{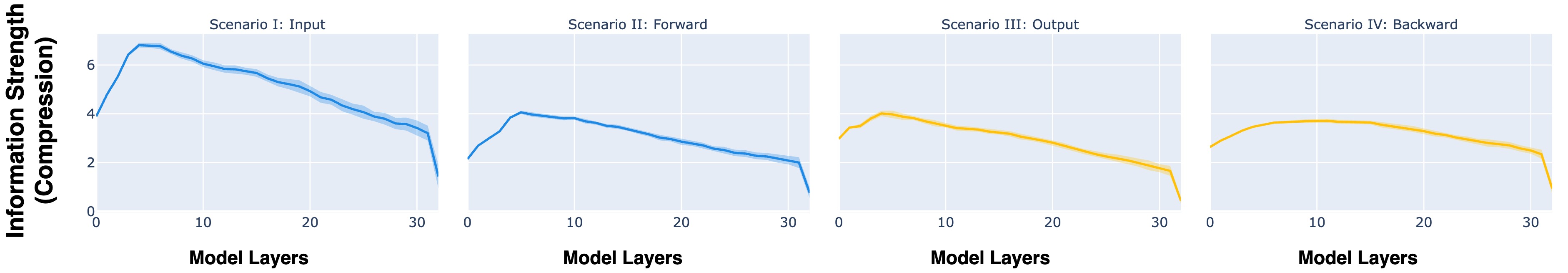}
    \caption{
    We further verify our probing setups and evaluate the compression of our probes~\citep{voita-titov-2020-information}, indicating how well information can be compressed.
    When compression is high, we assume strong patterns in the internal representations.
    These results show a similar trend to our results of an information peak in early layers, further justifying our probing setup.
    }
    \label{fig:compression}
\end{figure*}

\end{document}